\documentclass[journal]{IEEEtran}

\ifCLASSINFOpdf
\else
\fi

\usepackage{amsfonts}
\usepackage{amsmath}
\usepackage{graphicx}
\usepackage{hhline} 
\usepackage{makecell}
\usepackage{multirow}
\usepackage{booktabs}
\usepackage{subfig}
\usepackage[usenames,dvipsnames]{color}
\usepackage{colortbl}
\definecolor{mygray}{gray}{.9}

\begin{document}
%
\title{From W-Net to CDGAN: Bi-temporal Change Detection via Deep Learning Techniques}
%
%

\author{Bin~Hou,
        Qingjie~Liu$^*$,~\IEEEmembership{Member,~IEEE,}
        Heng~Wang,
        and~Yunhong~Wang,~\IEEEmembership{Fellow,~IEEE,}
        
\thanks{$^*$Correspondence to qingjie.liu@buaa.edu.cn.}
}

\markboth{IEEE Transactions on Geoscience and Remote Sensing,~Vol.~, No.~, March~2020}%
{Shell \MakeLowercase{\textit{et al.}}: Bare Demo of IEEEtran.cls for IEEE Journals}
%

\maketitle

\begin{abstract}

Traditional change detection methods usually follow the image differencing, change feature extraction and classification framework, and their performance is limited by such simple image domain differencing and also the hand-crafted features. Recently, the success of deep convolutional neural networks (CNNs) has widely spread across the whole field of computer vision for their powerful representation abilities. In this paper, we therefore address the remote sensing image change detection problem with deep learning techniques. We firstly propose an end-to-end dual-branch architecture, termed as the W-Net, with each branch taking as input one of the two bi-temporal images as in the traditional change detection models. In this way, CNN features with more powerful representative abilities can be obtained to boost the final detection performance. Also, W-Net performs differencing in the feature domain rather than in the traditional image domain, which greatly alleviates loss of useful information for determining the changes. 
Furthermore, by reformulating change detection as an image translation problem, we apply the recently popular Generative Adversarial Network (GAN) in which our W-Net serves as the Generator, leading to a new GAN architecture for change detection which we call CDGAN. To train our networks and also facilitate future research, we construct a large scale dataset by collecting images from Google Earth and provide carefully manually annotated ground truths. Experiments show that our proposed methods can provide fine-grained change detection results superior to the existing state-of-the-art baselines.
\end{abstract}


\IEEEpeerreviewmaketitle

\section{Introduction}

\IEEEPARstart{C}{hange} detection for remote sensing refers to identifying the differences between images acquired over the same geographical area at different times~\cite{tewkesbury2015critical}. It is widely used in disaster assessment~\cite{fan2017quantifying}, environmental monitoring~\cite{mucher2000land}, and urban expansion~\cite{xiao2006evaluating}, etc. With the development of earth observation techniques in recent years, an increasing number of high spatial and spectral resolution remote sensing images gain easier availability, which provides new opportunities and meanwhile challenges to change detection tasks with involvement of detailed spatial and contextual information. 

Most existing change detection methods follow a three-step workflow.  
Firstly, image preprocessing techniques, such as co-registration and denoising, are applied in order to make the images  as comparable as possible. Then, difference images (DIs) are generated by comparing multi-temporal images in either a pixel-wise or a segment-wise manner, with image differencing, image ratioing and their variations. Finally, change features are extracted from the DIs, and change maps are obtained by analyzing the change features through classification.

The last step is the core part of change detection methods. After obtaining features of the two input images,  
change detection is usually treated as an unsupervised or supervised classification problem. Unsupervised methods usually identify changes via thresholding~\cite{bruzzone2000minimum,bovolo2011adaptive} or clustering~\cite{Krinidis2010A,ghosh2011fuzzy} strategies. Though attractive in real applications for not requiring to collect and label ground truth data~\cite{bovolo2012framework,liu2015hierarchical},  these methods cannot provide detailed change information, thus give poor results when changed and unchanged features are overlapped or their statistical distributions  are modeled inaccurately. The performance would be even worse as the spatial resolution of images increases. Supervised classification~\cite{volpi2013supervised,li2015change} is a complementary solution to the unsupervised counterpart, which leverages prior information of land covers or changes. These methods are able to provide more accurate from-to information of changes. 
Either supervised or unsupervised, most existing methods rely on the hand-crafted feature representations, and consequently suffer from limited representative abilities to model complex and high level change information, leading to rather poor performance under clutter land covers.

High-level features are crucial for classification tasks because they are more robust and invariant to some distracting factors such as noise and scale variations. In this work, we propose to exploit deep learning techniques for change detection, which are well proved to be able to extract hierarchical features of input images straightforwardly.
Deep learning based architectures, specially CNNs, can effectively model the spatial context. CNNs are capable of learning complex geometric features by using non-linear activation functions in multilayer network configuration. They take as input a training image and outputs a single label for each image. But for change detection, there are two input images that need to be converted to one input for the network, which may lead to loss of useful information. Based on this consideration, we design an end-to-end dual-branch deep network for change detection, which is able to learn more robust and abstract representations for the neighborhood gray information of the given pixel. A fully convolutional network architecture including encoder and decoder is adopted due to its superiority for pixel-wise classification. With such a dual-branch architecture, we are able to realize the feature expression of each image, thus effectively avoid the information lose compared with using a single-branch architecture. The proposed network is termed the ``W-Net" and can produce a change map directly from two images by integrating the two processes of difference image (DI) generation and classification.

Furthermore, in contrast to the traditional DIs generation in the image domain, we conduct comparison in the feature domain and obtain difference information in a learning manner. We  generate change maps from the difference features with a decoder-like architecture. In the proposed network, the two branches accept bi-temporal images as input for feature extraction. Shortcut connections are used for  information propagation from low to high resolution layers. Features from the two branches are fused through concatenation during the network learning.

To further enhance feature representation and generalization abilities of the detection model, we also propose to introduce Generative Adversarial Networks (GANs) to solving change detection tasks. Traditional discriminant networks are heavily dependent on the statistical properties of the images and easily affected by unbalanced training data. Consequently, they can well perform in the same data domain but tend to fail to generalize to other related data domains.
In this work, we incorporate the above developed W-Net as Generator of a GAN architecture for detecting changes, and the traditional image classification task is therefore converted to an image translation problem.  We name such an architecture the ``CDGAN". 
Through the continuous adversarial learning, the network can establish a distribution connection between the two input images and DI, rather than just regarding change detection as the process of feature generation and classification.

In addition, to better facilitate implementation of the proposed methods, we construct a large scale dataset for change detection. It contains 29,000 pairs of remote sensing image patches with size of $256\times256$ pixels. These image patches are derived from 29 pairs of bi-temporal high resolution images collected with Google Earth and carefully annotated by some experts. 
It is expected to greatly alleviate the demand for large volumes of training samples of supervised, especially deep learning based change detection methods. We promise to make the dataset publicly available upon acceptance.

To summarize, we make the following  contributions:
\begin{itemize}

	\item[1] We propose W-Net for bi-temporal remote sensing image change detection, which is an end-to-end dual-branch network that accepts two images as input and extracts features from them independently, and generates change maps with the extracted difference features. 
	
	\item[2] We are among the first to reformulate change detection as an image translation problem and address it with conditional generative adversarial learning. Our CDGAN uses the proposed W-Net as Generator to construct the GAN architecture, and can accurately learn the distribution of change features and generate promising results.
	
	\item[3] A large and challenging dataset is collected and labeled for change detection, which would greatly alleviate the demand of massive training samples for deep learning based change detection methods. 
\end{itemize}

\section{Related Work}\label{related work}

\subsection{Traditional Change Detection}

Traditional change detection approaches can be grouped into pixel-based and object-based methods according to the unit of image analysis. Pixel-based methods make comparisons of bi-temporal images and extract features from them in a pixel-wise manner. In addition to the simple algebra methods, such as image differencing, image ratioing, and image regression,
the most widely used technique is change vector analysis (CVA)~\cite{malila1980change,bovolo2007theoretical,chen2016multi}, which first computes a multispectral difference image and then uses both magnitude and direction of difference vectors in a two-dimensional space for identifying changes. These methods 
do not consider local information, thus are sensitive to noise and misregistration errors. One solution is transforming the raw feature vectors into a new feature space to reduce noise impact. Principal component analysis (PCA)~\cite{deng2008pca}, iterative reweighted multivariate alteration detection (IR-MAD)~\cite{nielsen2007regularized} and wavelet transform~\cite{celik2010unsupervised} are popular image transformation methods. Another solution is incorporating spatial and contextual information of pixels by modeling local relationships of pixels~\cite{volpi2013supervised,xiong2012threshold,gu2017change,falco2013change}.
Generally, the pixel-based methods are more applicable to low and medium resolution remote sensing images than high resolution ones.

Comparatively, object-based change detection (OBCD) methods compare and analyze meaningful objects with regular shapes and sizes obtained from segmentation methods. For example, Im et al.~\cite{im2008object} proposed an object-based change detection method based on object/neighborhood correlation image analysis and image segmentation. Bovolo~\cite{bovolo2009multilevel} proposed a novel parcel-based context-sensitive change detection technique for very high resolution (VHR) remote sensing images, which models scenes at different resolutions by defining multitemporal and multilevel parcels and exploiting the spatial-context information. Huo et al.~\cite{huo2010fast} 
improved the discriminability between changed and unchanged classes using object-level features and boosted performance using progressive change feature classification. Chen et al.~\cite{Jianyu2013Detecting} presented a spatial contrast-enhanced image object-based change detection approach (SICA), in which image object detection approach (IODA) is used to identify changed areas in high-resolution satellite images by integrating shape changes into traditional change detection. Robertson and King~\cite{King2011Comparison} compared pixel- and object-based classification in landcover change mapping and revealed that the object-based approach could depict change information more accurately.

\subsection{Deep Learning Based Change Detection}

Deep learning techniques have gained wide success in many vision tasks in recent years\cite{zhang2018road,Wang2018Scene,liu2018remote}. CNNs are among the most popular deep architectures, and often used as backbone or feature extractors to solve vision tasks for their good generalizability, such as AlexNet~\cite{krizhevsky2012imagenet}, VGGNet~\cite{simonyan2014very}, and GoogleNet~\cite{szegedyrabinovich}.

Some researchers harness the capabilities of CNNs as well as other deep learning methods for remote sensing applications. For instance, Gong et al.~\cite{Gong2017Change} designed an end-to-end deep neural network using a stack of restricted Boltzmann machine (RBM) and back propagation (BP) to produce change detection maps directly. Gao et al.~\cite{Gao2016Automatic} proposed a semi-supervised change detection model for synthetic aperture radar (SAR) images based on PCANet, in which training samples are obtained using Gabor wavelets and fuzzy c-means. Zhang et al.~\cite{Zhang2016DeepandMapping} presented a novel multi-spatial-resolution change detection framework incorporating deep-architecture-based unsupervised feature learning and mapping-based feature change analysis. Zhang et al.~\cite{Zhang2016FeaturelevelDeep} combined deep belief networks (DBNs) and  feature change analysis based on cosine angle distance (CAD) to highlight the changes, and then mapped bitemporal change features into a 2D polar domain for final classification. The above architectures are relatively simple, mainly including a stack of fully connected layers, thus offer limited abilities to extract discriminative features. Most of these works adopt the handcrafted features and traditional algorithms for training sets production and simple classification strategies for binary change map generation.

To avoid shortcomings of 1D neural networks, 2D CNNs are also used for change detection. Liu et al.~\cite{Liu2016A} proposed a symmetric convolutional coupling network (SCCN) for heterogeneous images by transforming the features into a consistent feature space. Hou et al.~\cite{hou2017change} developed a low rank and deep feature based change detection method, using fine-tuned VGGNet to extract hypercolumn features of the objects. Zhan et al.~\cite{Zhan2017Change} trained a siamese CNN using the weighted contrastive loss to make unchanged pixels have similar features and changed ones have distinct features. Wang et al.~\cite{Wang2018GETNET} presented an end-to-end 2D CNN framework for hyperspectral image change detection, where mixed-affinity matrices are formed from which GETNET extracts features for classification. Zhang et al.~\cite{Wu2019SpectralSpatial} proposed a spectral-spatial joint learning network (SSJLN), using a network similar to the siamese CNN to extract spectral-spatial joint representations and fuse them to represent difference information, and using the discrimination learning to explore the underlying information. Ma et al.~\cite{Ma2019DeepCapsule} proposed a change detection method for heterogeneous images based on pixel-level mapping and a capsule network with a deep structure. Though these methods adaptively learn spectral and spatial information by using 2D CNNs, they perform feature extraction of two input images respectively, without considering the information connectivity between them. In addition, they obtain difference images by some traditional feature fusion strategies. Comparatively, our proposed network can automatically learn the change features and perform final classification with the interconnected dual-branch fully convolutional neural network structure.

Recently, Generative Adversarial Networks (GANs)~\cite{Goodfellow2014Generative} have gained much popularity for their good capability of generating high-quality images~\cite{ledig2016photo,Isola_2017_CVPR,liu2018psgan}. 
Observing DIs are the key to performance of change detection, Gong et al.~\cite{Gong2017Generative} modeled the distributions of training data and produced better DIs based on a GAN architecture.
For generator and discriminator in their method, the fully-connected layer is also used for feature extraction or classification; a fuzzy local information c-means clustering algorithm (FLICM)~\cite{Krinidis2010A} is applied to generate final change maps. Gong et al. \cite{Gong2019GenerativeDiscriminatoryClassified} also proposed a generative discriminatory classified network (GDCN) for multispectral image change detection. Its generator converts the input noise into fake data that match real image data, and then labeled data and unlabeled data obtained by preclassification, as well as fake data are all input into DCN, leading to a GDCN architecture.
Niu et al. \cite{Niu2019ConditionalAdversarialNetwork} adopted a conditional generative adversarial network (cGAN) for change detection over heterogeneous images.  It contains a cGAN-based translation network aiming to translate the optical image with the SAR image as a target, and an approximation network that approximates the SAR image to the translated one by reducing their pixelwise difference. However, their used network architectures all include simple fully connected layers, taking a pixel or patch as input. In these networks, the neighborhood and contextual information around a pixel is not properly considered, and they suffer from a huge amount of learnable parameters. In addition, the selection of training samples depends on naive existing methods, and consequently the subsequent process may get harmed by resultant errors. Our proposed CDGAN also uses fully convolutional layers as generator and discriminator for adaptively modeling the geographic objects distribution, but our network training relies on lots of manually annotated data, which provides reliable data assurance. Then change detection performance is improved via realizing image generation.

\section{Proposed Methods}\label{approach}

In this section, we first elaborate on details of the proposed W-Net for change detection. Then, we explain how we apply adversarial training to detecting changes in the images based on the proposed W-Net, leading to a new GAN architecture named CDGAN.

\begin{figure*}[!ht]
	\centerline{\includegraphics[width=0.6\textwidth]{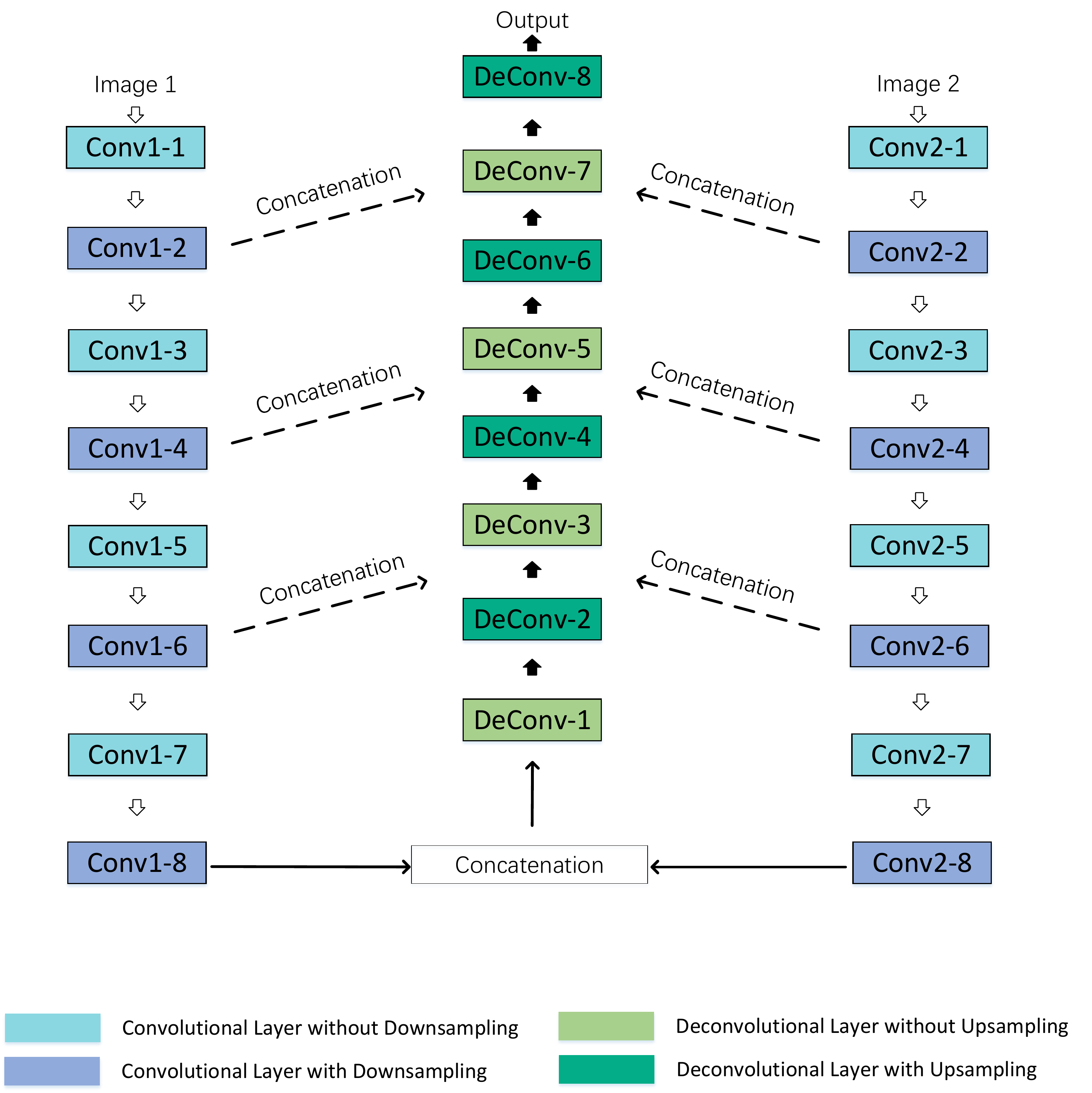}}
	\caption{Architecture of the proposed W-Net. W-Net includes encoder part and decoder part. The encoder contains four convolutional blocks, and each block consists of one traditional convolutional layer with stride 1 and one strided convolutional layer with stride 2. The decoder contains four deconvolutional blocks, with each consisting of one deconvolutional layer with stride 1 and one deconvolutional layer with stride 2. There are three shortcut connections between encoder and decoder.}
	\label{fig:w-net}
\end{figure*}

\begin{table*}[htb]
	\tiny
	\centering
	\caption[justification=centering]{The detailed parameters of the W-Net.}
	\resizebox{0.7\textwidth}{!}{ %
		\begin{tabular}{cccc}
			\Xhline{0.5pt}
			\toprule
			
			
			Layer Name & Kernel Size & Stride & Output Size \\ 
			\cmidrule(lr){1-4} 
			
			Input& - & - & $256\times256\times3$\\
			\cmidrule(lr){1-4}

			Conv1-1,Conv2-1& $3\times3$ & 1 & $256\times256\times64$ \\
			\cmidrule(lr){1-4}

			Conv1-2,Conv2-2& $3\times3$ & 2 & $128\times128\times128$ \\
%
%
			\cmidrule(lr){1-4}

			Conv1-3,Conv2-3& $3\times3$ & 1 & $128\times128\times256$ \\
			
			\cmidrule(lr){1-4}

			Conv1-4,Conv2-4& $3\times3$ & 2 & $64\times64\times512$ \\
			
%
%
			\cmidrule(lr){1-4}

			Conv1-5,Conv2-5& $3\times3$ & 1 & $64\times64\times512$ \\
			
			\cmidrule(lr){1-4}

			Conv1-6,Conv2-6& $3\times3$ & 2 & $32\times32\times512$ \\
			
%
%
			\cmidrule(lr){1-4}

			Conv1-7,Conv2-7& $3\times3$ & 1 & $32\times32\times512$ \\
			
			\cmidrule(lr){1-4}

			Conv1-8,Conv2-8& $3\times3$ & 2 & $16\times16\times(512+512)$ \\
			
			\cmidrule(lr){1-4}

			DeConv-1& $3\times3$ & 1 & $16\times16\times512$ \\
			
			\cmidrule(lr){1-4}

			DeConv-2& $3\times3$ & 2 & $32\times32\times(512+512+512)$ \\
			
			\cmidrule(lr){1-4}

			DeConv-3& $3\times3$ & 1 & $32\times32\times512$ \\
			
			\cmidrule(lr){1-4}

			DeConv-4& $3\times3$ & 2 & $64\times64\times(512+512+512)$ \\
			
			\cmidrule(lr){1-4}

			DeConv-5& $3\times3$ & 1 & $64\times64\times256$ \\
			
			\cmidrule(lr){1-4}

			DeConv-6& $3\times3$ & 2 & $128\times128\times(128+128+128)$ \\
			
			\cmidrule(lr){1-4}

			DeConv-7& $3\times3$ & 1 & $128\times128\times64$ \\
			
			\cmidrule(lr){1-4}

			DeConv-8& $3\times3$ & 2 & $256\times256\times1$ \\

			\bottomrule
			\Xhline{0.5pt}
	\end{tabular}}%
	\label{tab:exp-wnet-architecture}
\end{table*}

\subsection{W-Net for Change Detection}\label{subsec: w-net}
\subsubsection{Formulation}
Given two images $X^{t_1}$ and $X^{t_2}$ captured at different time $t_1$ and $t_2$ over the same site and ground truth $CM$, we aim to identify change areas between these two images. Suppose $\hat{CM}$ is the change map inferred from $X^{t_1}$ and $X^{t_2}$, and $\hat{CM}_{i,j}$ is the change values at location $(i,j)$. Normally $\hat{CM}_{i,j} \in \{0,1\}$, $\hat{CM}_{i,j} = 1$ indicates $(i,j)$ is changed, and otherwise it indicates $(i,j)$ is unchanged. Instead of classifying $(i,j)$ as binary change values, we can estimate the probability of $(i,j)$ being changed by formulating it as a dense prediction problem:
\begin{equation}\label{eq:formulation}
    \hat{CM}_{i,j}=p(c_{i,j}=1|X^{t_1},X^{t_2};\Theta)
\end{equation}
where $c_{i,j}=1$ indicates pixel $(i,j)$ is changed, and $\Theta$ is the parameter set of the model, which can be solved by maximizing
\begin{equation}\label{eq:solve}
\Theta^{*} =\arg\max_{\Theta} \prod\limits_{i,j}  p(c_{i,j}=CM_{i,j}|X^{t_1},X^{t_2};\Theta).
\end{equation}
The change detection task can be modeled in Eqn.~(\ref{eq:formulation}) and solved by Eqn.~(\ref{eq:solve}) with back propagation.

\subsubsection{Architecture}
Our architecture design is based on following considerations. First, downsampling feature maps with strided convolution leads to better performance in low level vision tasks than with pooling~\cite{Isola_2017_CVPR,radford2015unsupervised}, thus we adopt strided convolution in our network. 
Secondly, in traditional change detection, the input two images need to be converted to one input for the single branch network, which would cause information loss, thus we use a dual-branch network each taking as input one of these two images. Such a network can also converge faster and more stably than a single branch one. Thirdly, we use the most common kernel size, $3\times3$. Though larger kernel size often indicates better performance due to larger receptive field, the number of parameters also increases drastically, making it harder to train. 

The proposed network consists of two branches, each of which is a sub-network extracting features from an input image. Each feature extraction network (short for FEN) consists of four convolutional blocks, and each block contains one $3\times3$ convolutional layer and one $3\times3$ strided convolutional layer for downsampling. In addition, the shortcut connection is added to each convolutional block for the transmission between low-level spectral information and high-level semantic information. We use $f^{t_{1}}_l$, $f^{t_{2}}_l \in \mathbb{R}^{w_l\times h_l\times d_l}$, $l = \{1,...,.L\}$, to represent feature maps in the  $l$-th layer of FEN at time $t_{1}$ and $t_{2}$. Finally, we obtain the features $f^{t_{1}}_L$ and $f^{t_{2}}_L$ for $X^{t_1}$ and $X^{t_2}$, respectively.

For high resolution remote sensing, high intra-class variability and low inter-class variability may lead to more information loss. The commonly used difference strategy would further aggravate this disadvantage. Comparatively, concatenation is more promising~\cite{Ding2015Sparse} and is a better choice for us. After obtaining features, change information is inferred from $f^{t_{1}}_L$ and $f^{t_{2}}_L$. We then concatenate the two features $(f^{t_{1}}_2 , f^{t_{2}}_2)$, $(f^{t_{1}}_4 , f^{t_{2}}_4)$, $\cdots$, $(f^{t_{1}}_L , f^{t_{2}}_L)$ from the outputs in the last convolutional layer of each convolutional block to form a joint feature representation of the two images, and learn change information from them to obtain the difference image. After that, a subsequent decoder-like network is used to predict the probabilities of changes in each pixel. The decoder is comprised of four deconvolutional blocks, each containing a $3\times3$ deconvolutional layer with stride 1 and a $3\times3$ deconvolutional layer with stride 2 for upsampling.   We apply rectified linear unit (ReLU) as the activation function after each convolutional layer and deconvolutional layer except for the last deconvolutional layer. With such  convolutional and deconvolutional architectures, the final output would have the same size with the inputs. The devised network is shaped like the letter $W$, so we call it W-Net. The architecture and detailed parameters of W-Net are illustrated in Fig. \ref{fig:w-net} and Table \ref{tab:exp-wnet-architecture}.

\subsubsection{Loss Function}
Given a set of training images and the corresponding change maps, our goal is to optimize the network through minimizing loss functions such that it can produce accurate change maps for the input bi-temporal images. In this work, we use common cross entropy as loss function to train the W-Net, defined as
\begin{equation}
	\mathcal{L}=-\frac{1}{N}\sum_{i,j}^N\sum_{c_{i,j}=0}^{1}y_{c_{i,j}}\log \hat{y}_{c_{i,j}},
\end{equation}
where $\mathcal{L}$ quantifies the misclassification by comparing the target label vectors $y$ with the predicted label vectors $\hat{y}$, and $N$ means the number of training samples.

\begin{figure*}[ht]
	\centerline{\includegraphics[width=1\textwidth]{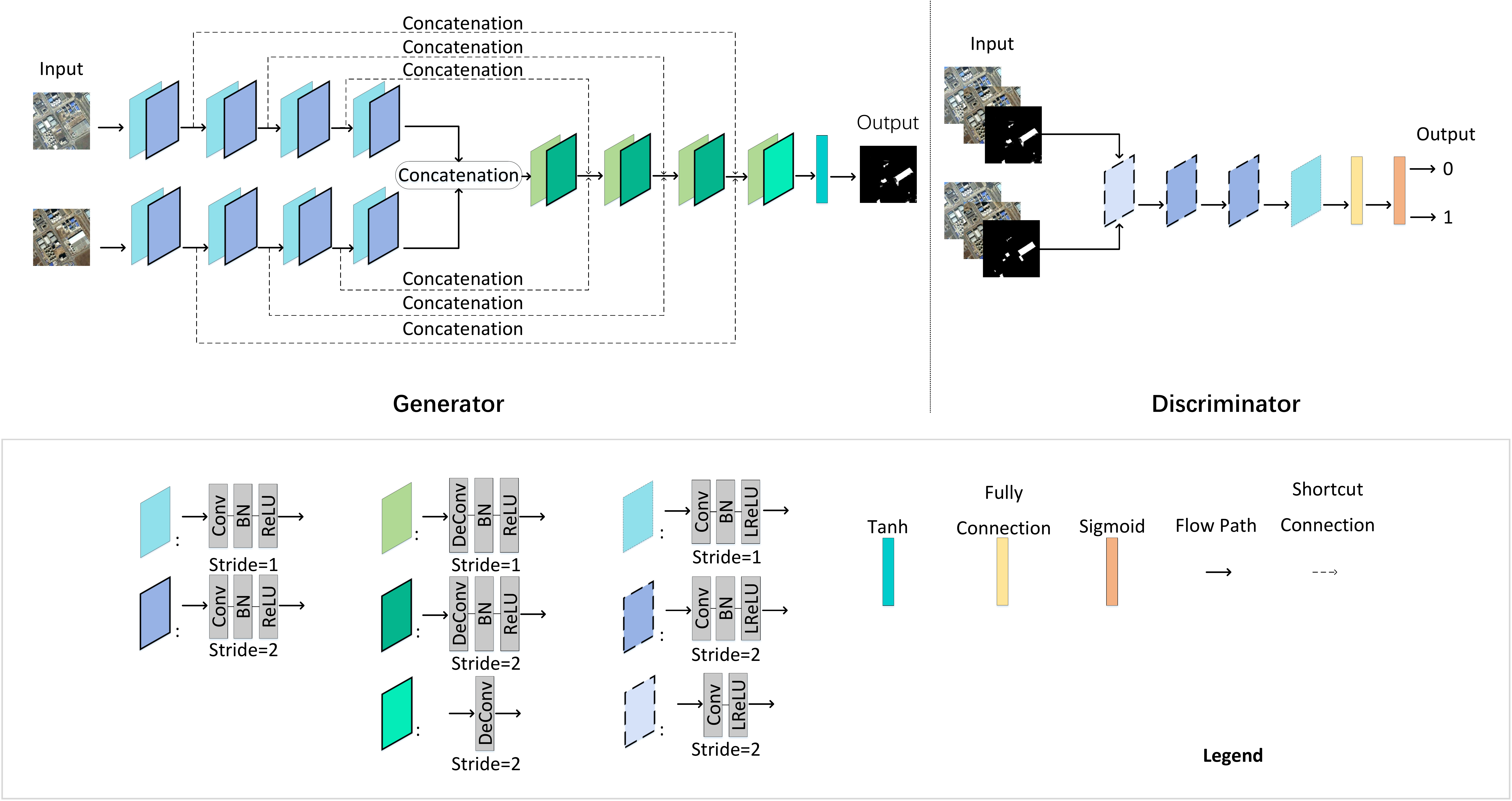}}
	\caption{Architecture of CDGAN. The generator of CDGAN is the proposed W-Net in Section~\ref{subsec: w-net}, which accepts the two images at different times as input and outputs the change map. The discriminator is a simple deep convolutional network in Section~\ref{subsec: gan}, with the two images at different times, and the change map generated by generator or ground truth as input, and 1 or 0 standing for changed or unchanged class as output.}
	\label{fig:flowchart}
\end{figure*}

\subsection{GAN Based Change Detection (CDGAN)}\label{subsec: gan}

\subsubsection{Formulation}
The traditional GANs learn a mapping from a random noise vector $z \sim p_z(z)$ to the output image $CM$. Given two input images $X^{t_1}$ and $X^{t_2}$ and an input noise variable $z$, the proposed GAN based change detection method, which we term CDGAN, can be modeled as inferring the change maps from the joint distribution of $p(X^{t_1}, X^{t_2}, z)$ instead, i.e., to learn a mapping function $G$: $(X^{t_1}, X^{t_2}, z)\rightarrow CM$. It can be rewritten as a conditional density estimation model $p(CM|X^{t_1}, X^{t_2}, z)$. Inspired by~\cite{Isola_2017_CVPR}, it can be estimated using the conditional GAN framework, which is comprised of a generator network $G$ and a discriminator network $D$. The generator $G$ with parameter $\theta_g$ is optimized to map $X^{t_1}$, $X^{t_2}$ and $z$
to the data space $\hat{CM}=G(X^{t_1}, X^{t_2}, z; \theta_g)$. The discriminator $D$ with parameter $\theta_d$ is used to output the probability $D(X^{t_1}, X^{t_2}, CM; \theta_d)$ where $CM$ is from the real data distribution $p_{data}(X^{t_1}, X^{t_2},CM)$, and the probability $D(X^{t_1}, X^{t_2}, \hat{CM}; \theta_d)$ is derived from the generator $G$.

\subsubsection{Architecture}
We adopt the above presented W-Net as the generator of the CDGAN. For better training and more stable convergence, we adopt kernel size $5\times5$ for both generator and discriminator. The ReLU activation is also used in the generator except for the output layer which uses the Tanh function. For the discriminator, the network is a conventional convolutional structure composed of four $5\times5$ convolutional layers. The last convolutional layer is flattened for a fully connected layer and then fed into a single sigmoid output. See Fig. \ref{fig:flowchart} for an illustration of the proposed CDGAN architecture. The convolutional layers all use leaky ReLU activations for better effects, except for the final fully connected layer, which uses a sigmoid activation.

\subsubsection{Loss Function}\label{subsec:loss}
The learning objective for change detection based on GAN corresponds to a minimax two-player game, which is formulated as

\begin{equation}\label{equ-objective-discriminator}
\begin{split}
&\min_G\max_D\mathcal{L}(G, D) =\\ 
&\mathbb{E}_{x,y\sim p_{data}(x,y)}[\log D(x,y)]+\\
&\mathbb{E}_{x\sim p_{data}(x),z\sim p_z(z)}[\log(1-D(x,G(x,z))),
\end{split}     
\end{equation}
where the generator network $G$ generates samples from an observed image $x$ and a random noise $z$, and the discriminator network $D$ is trained to distinguish whether a sample belongs to the real data $y$ or is generated by $G$. A combination of the error from the discriminator and the $L1$ distance w.r.t. the ground truth can be used to improve the stability and convergence rate of the adversarial training in the generator as shown in Eqn. (\ref{equ-objective-generator}):
\begin{equation}\label{equ-objective-generator}
\begin{split}
G^*&=\arg\min_G\max_D\mathcal{L}(G,D)+\lambda\mathcal{L}_{L1}(G)\\
\text{and }\mathcal{L}_{L1}(G)&=\mathbb{E}_{x,y\sim p_{data}(x,y),z\sim p_z(z)}[\|y-G(x,z)\|_1].
\end{split}
\end{equation}

\begin{figure*}[htb]
	\hspace{0.4cm}
	\begin{minipage}[b]{.2\linewidth}
		\centering
		\centerline{\includegraphics[width=4.4cm]{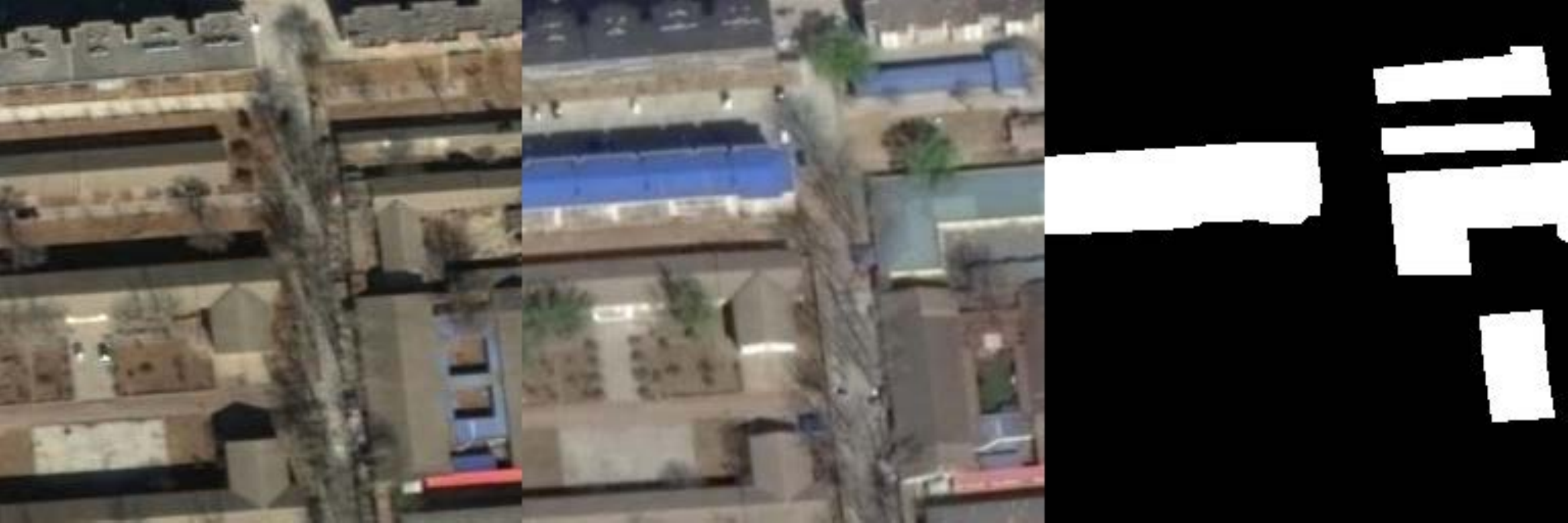}}
		\vspace{0.1cm}
	\end{minipage}
	\hfill
	\begin{minipage}[b]{.2\linewidth}
		\centering
		\centerline{\includegraphics[width=4.4cm]{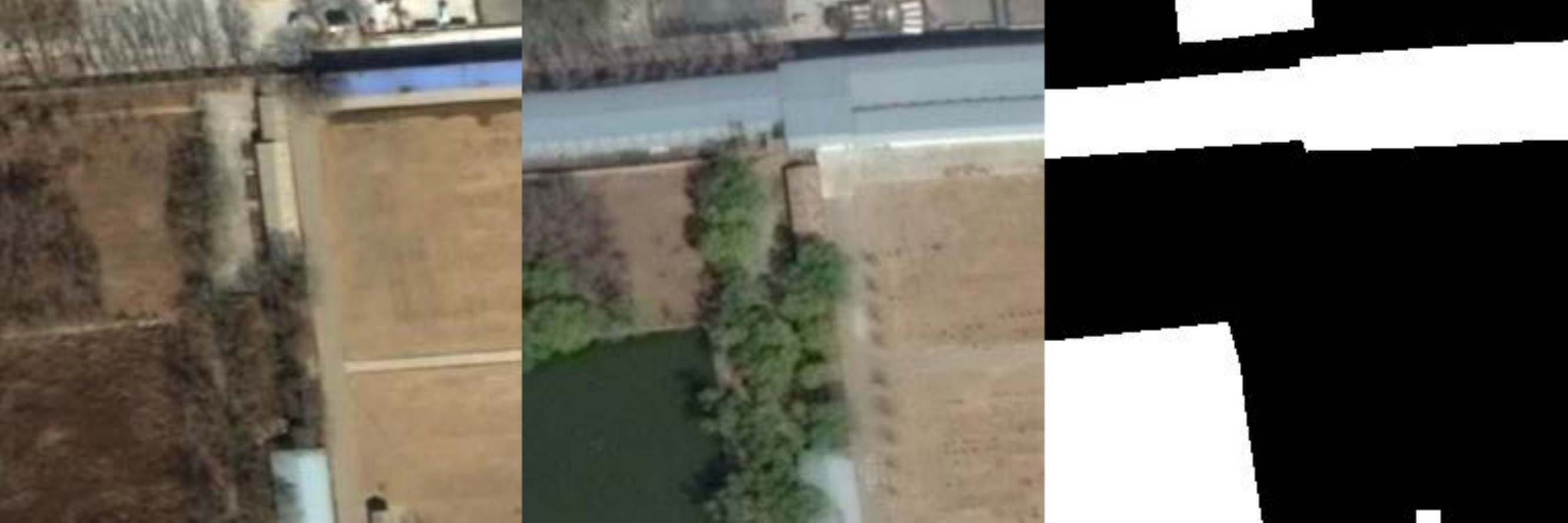}}
		\vspace{0.1cm}
	\end{minipage}
	\hfill
	\begin{minipage}[b]{.2\linewidth}
		\centering
		\centerline{\includegraphics[width=4.4cm]{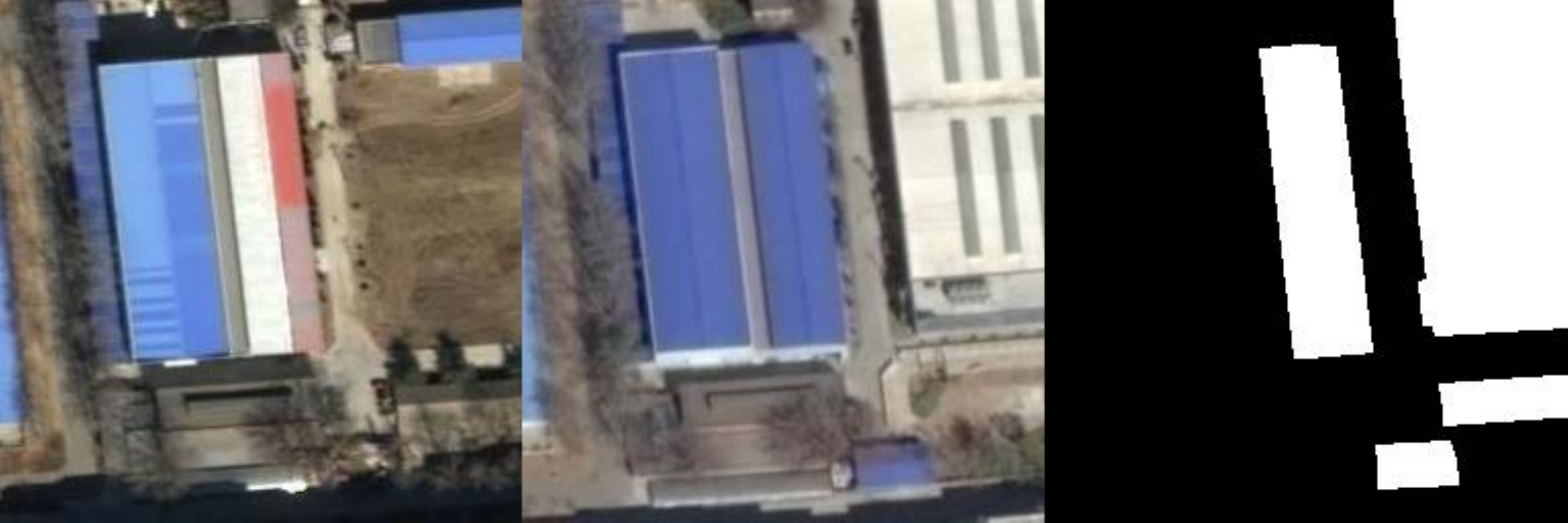}}
		\vspace{0.1cm}
	\end{minipage}
	\hfill
	\begin{minipage}[b]{.2\linewidth}
		\centering
		\centerline{\includegraphics[width=4.4cm]{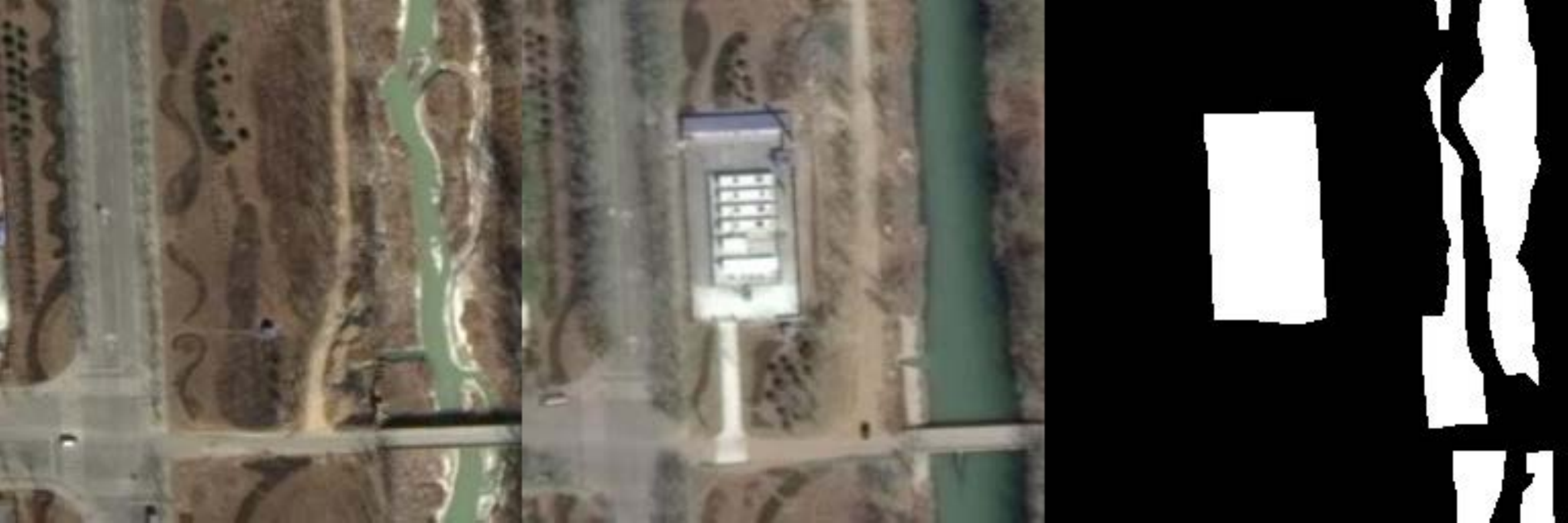}}
		\vspace{0.1cm}
	\end{minipage}
	\hspace{0.4cm}
	
	\vfill
	
	\hspace{0.4cm}
	\begin{minipage}[b]{.2\linewidth}
		\centering
		\centerline{\includegraphics[width=4.4cm]{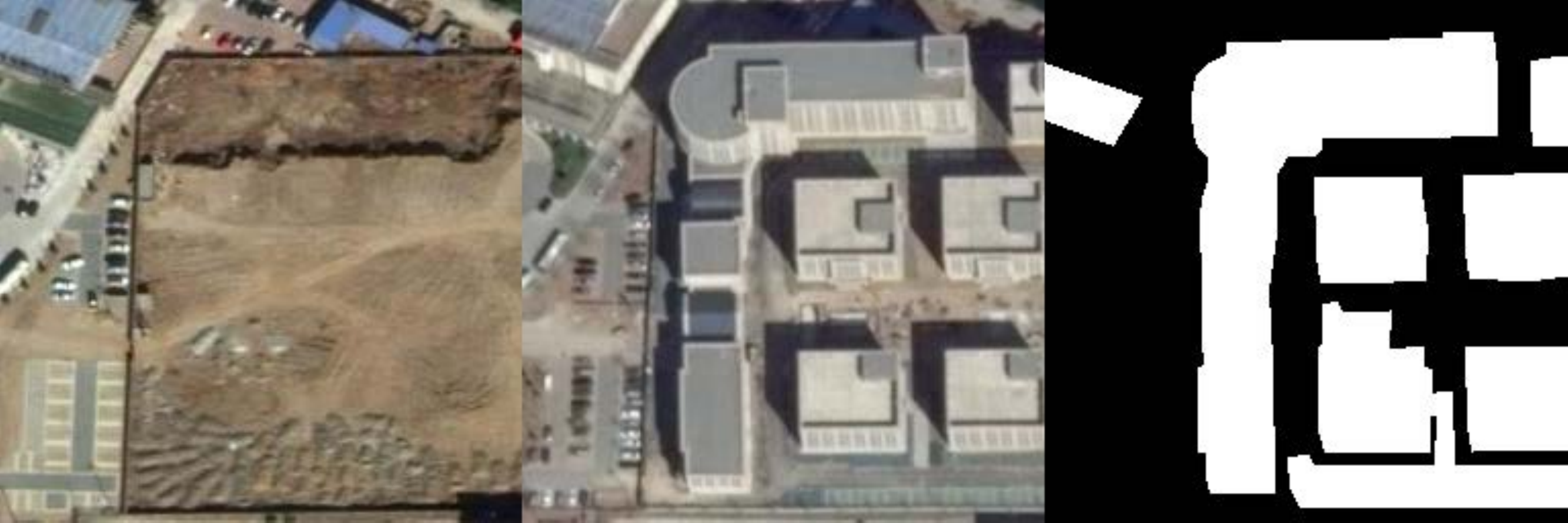}}
		\vspace{0.1cm}
	\end{minipage}
	\hfill
	\begin{minipage}[b]{.2\linewidth}
		\centering
		\centerline{\includegraphics[width=4.4cm]{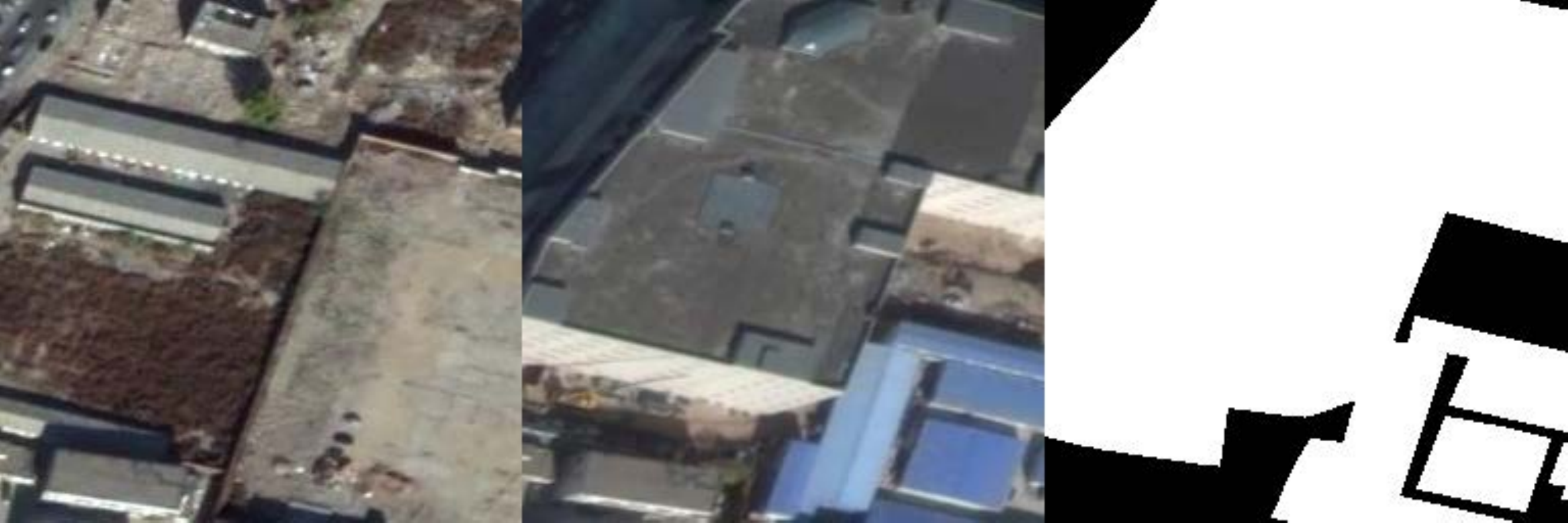}}
		\vspace{0.1cm}
	\end{minipage}
	\hfill
	\begin{minipage}[b]{.2\linewidth}
		\centering
		\centerline{\includegraphics[width=4.4cm]{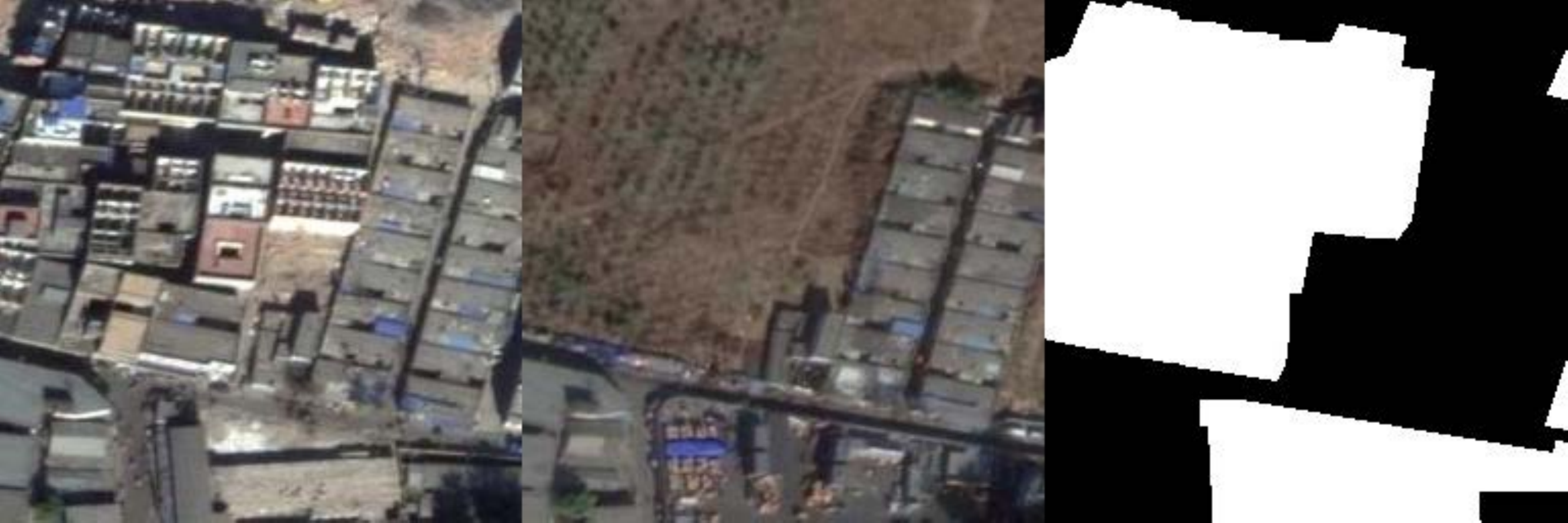}}
		\vspace{0.1cm}
	\end{minipage}
	\hfill
	\begin{minipage}[b]{.2\linewidth}
		\centering
		\centerline{\includegraphics[width=4.4cm]{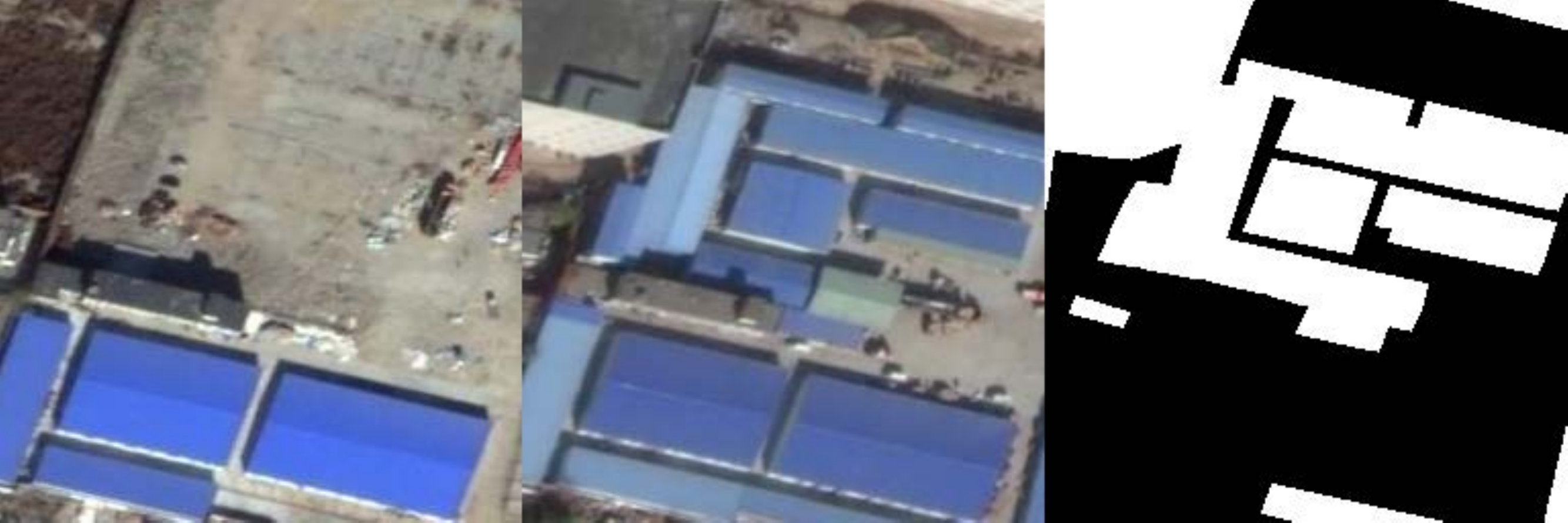}}
		\vspace{0.1cm}
	\end{minipage}
	\hspace{0.4cm}
	
	\vfill
	
	\hspace{0.4cm}
	\begin{minipage}[b]{0.2\linewidth}
		\centering
		\centerline{\includegraphics[width=4.4cm]{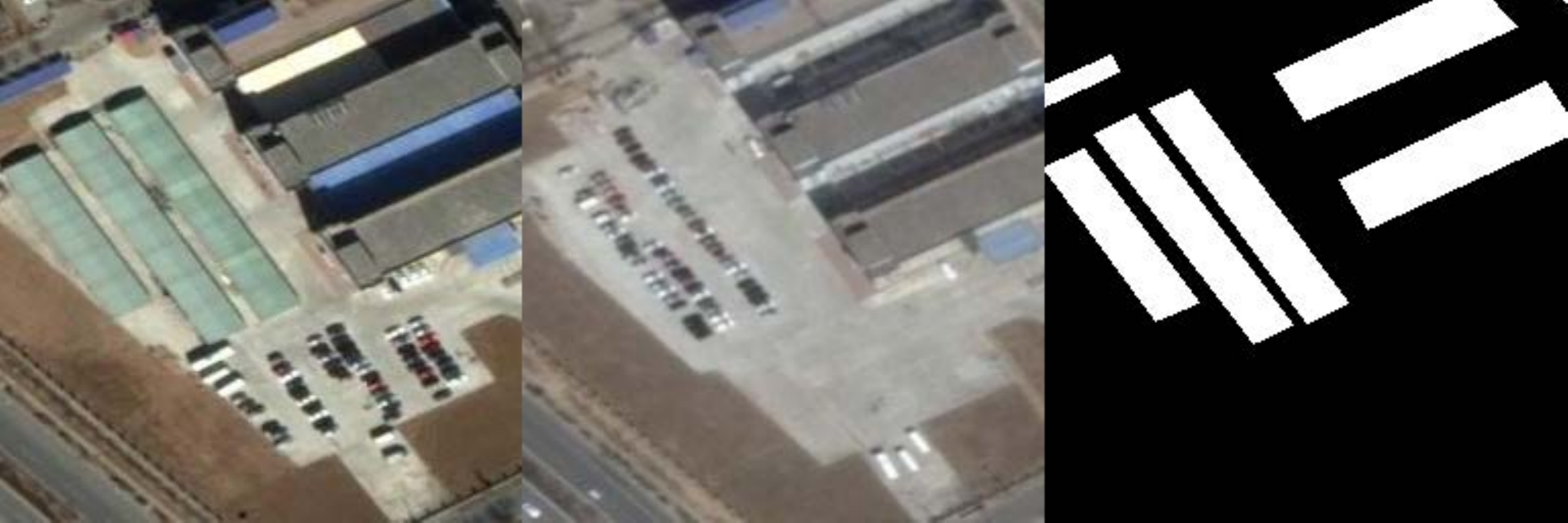}}
		\vspace{0.1cm}
	\end{minipage}
	\hfill
	\begin{minipage}[b]{0.2\linewidth}
		\centering
		\centerline{\includegraphics[width=4.4cm]{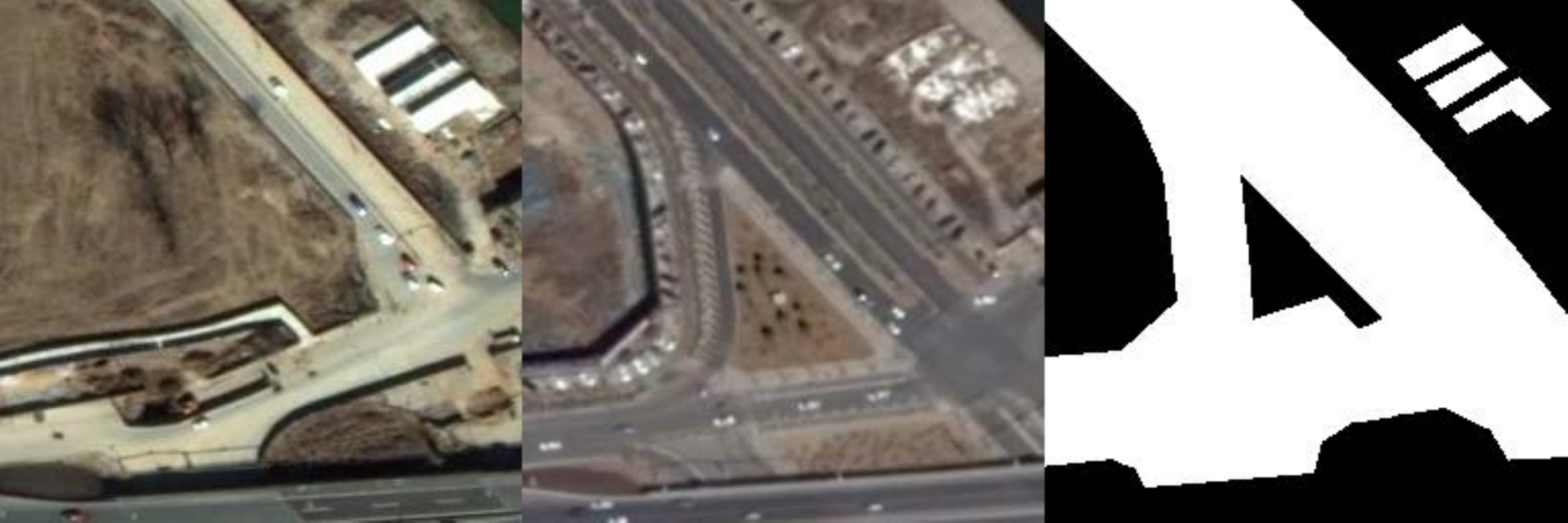}}
		\vspace{0.1cm}
	\end{minipage}
	\hfill
	\begin{minipage}[b]{0.2\linewidth}
		\centering
		\centerline{\includegraphics[width=4.4cm]{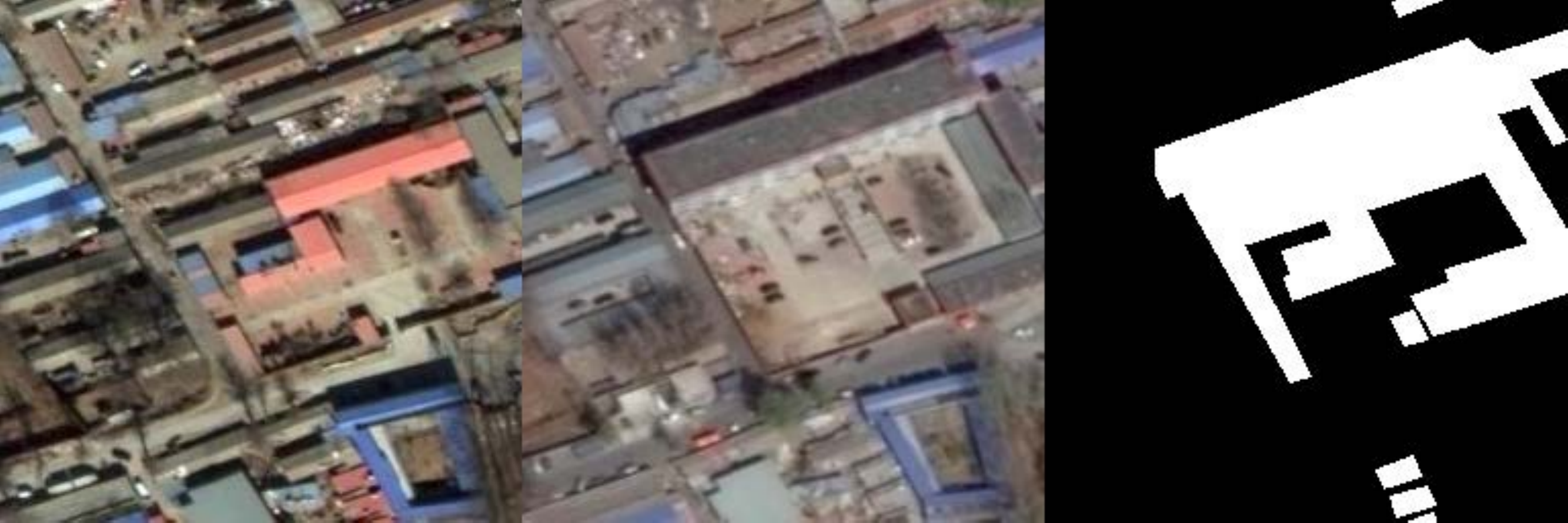}}
		\vspace{0.1cm}
	\end{minipage}
	\hfill
	\begin{minipage}[b]{0.2\linewidth}
		\centering
		\centerline{\includegraphics[width=4.4cm]{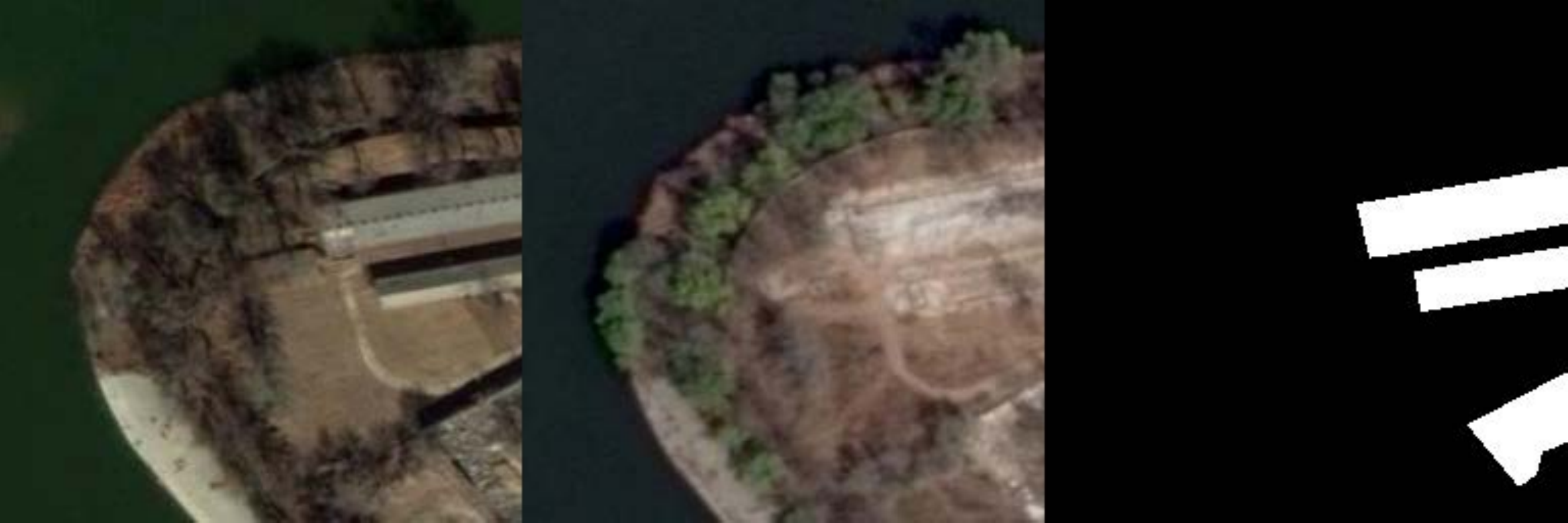}}
		\vspace{0.1cm}
	\end{minipage}
	\hspace{0.4cm}
	
	\vfill
	
	\hspace{0.4cm}
	\begin{minipage}[b]{0.2\linewidth}
		\centering
		\centerline{\includegraphics[width=4.4cm]{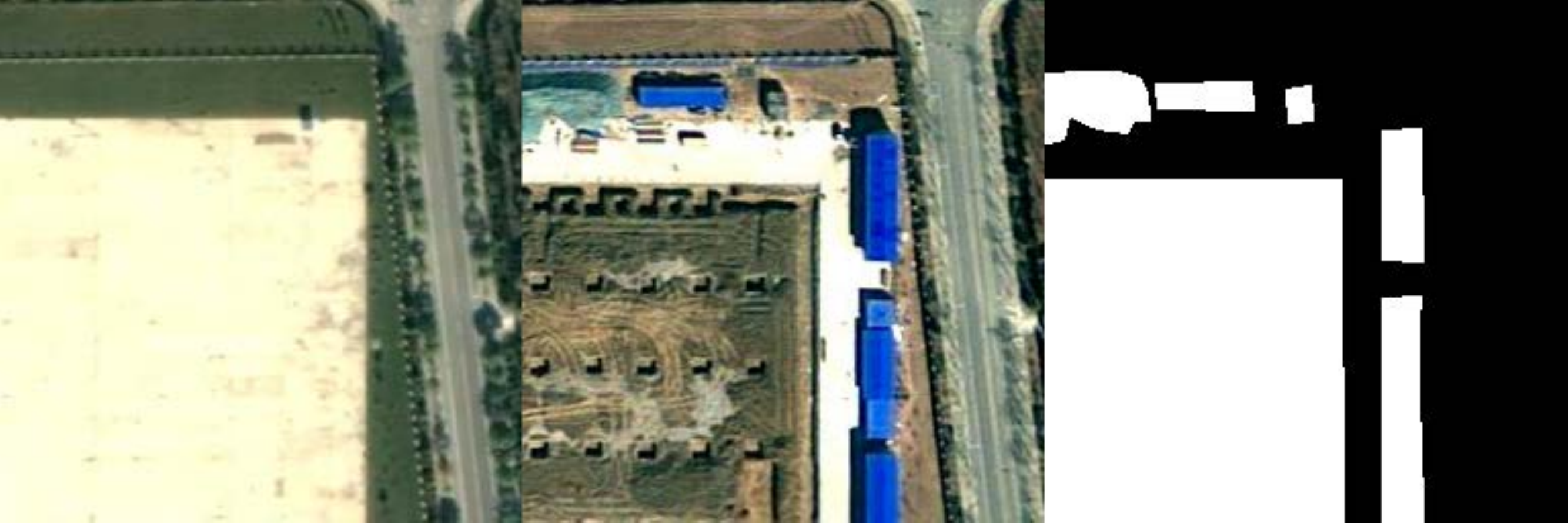}}
		\vspace{0.1cm}
	\end{minipage}
	\hfill
	\begin{minipage}[b]{0.2\linewidth}
		\centering
		\centerline{\includegraphics[width=4.4cm]{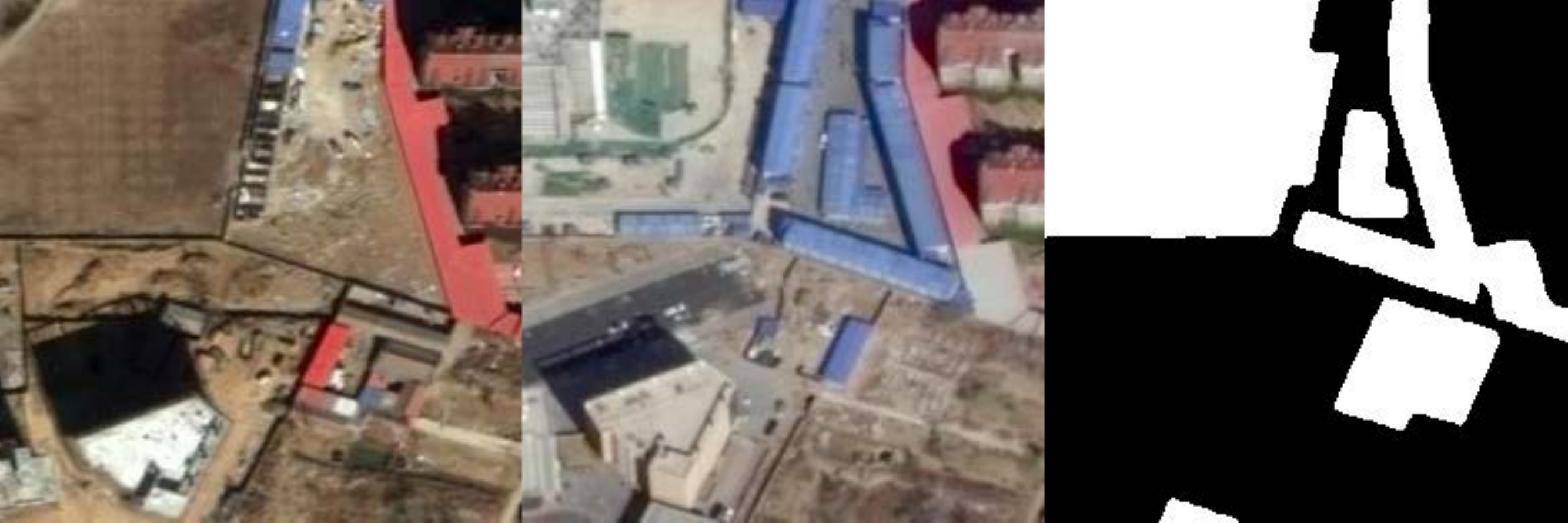}}
		\vspace{0.1cm}
	\end{minipage}
	\hfill
	\begin{minipage}[b]{0.2\linewidth}
		\centering
		\centerline{\includegraphics[width=4.4cm]{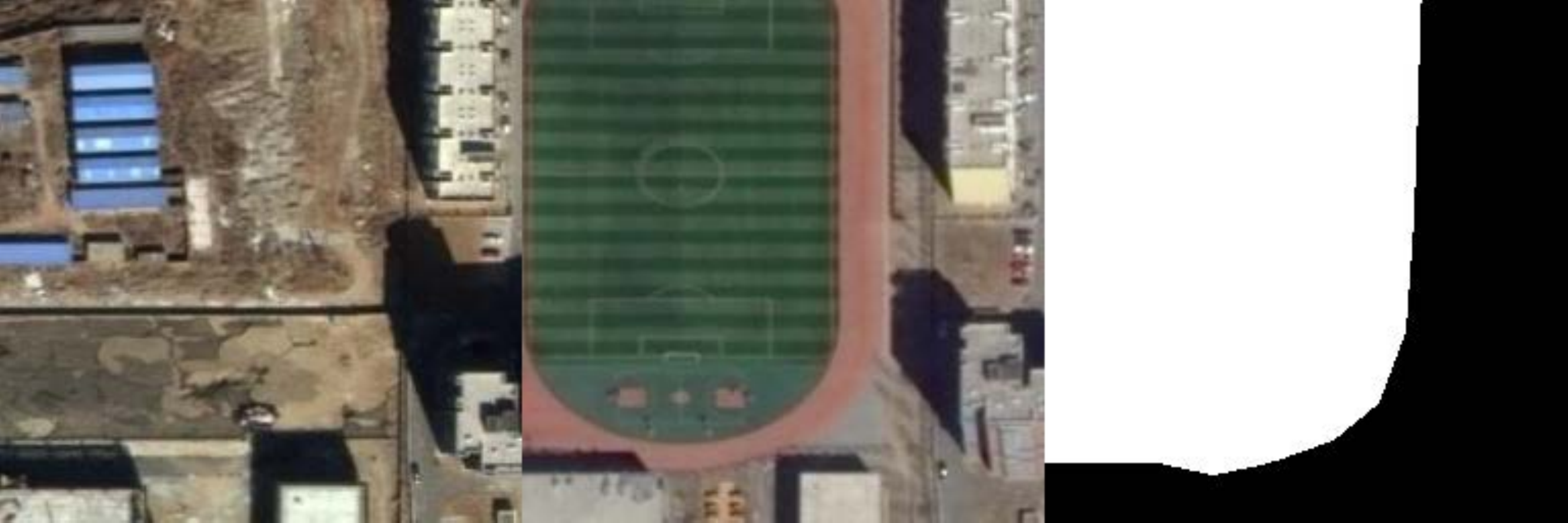}}
		\vspace{0.1cm}
	\end{minipage}
	\hfill
	\begin{minipage}[b]{0.2\linewidth}
		\centering
		\centerline{\includegraphics[width=4.4cm]{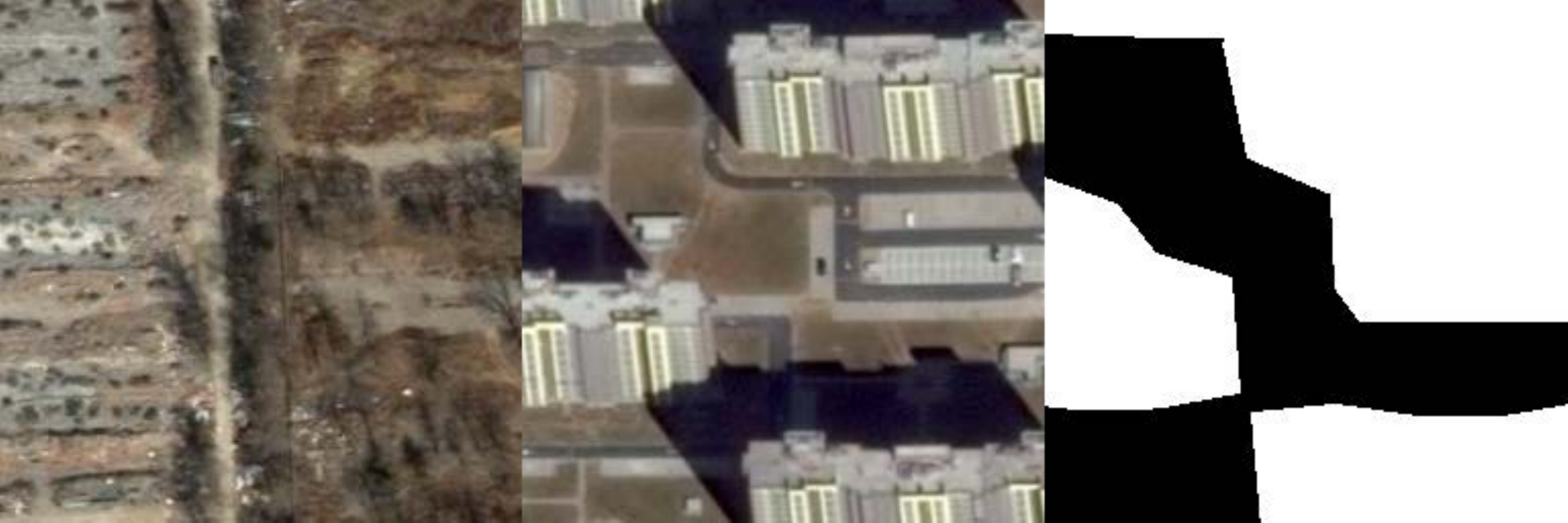}}
		\vspace{0.1cm}
	\end{minipage}
	\hspace{0.4cm}
	
	\vfill
	
	\hspace{0.4cm}
	\begin{minipage}[b]{0.2\linewidth}
		\centering
		\centerline{\includegraphics[width=4.4cm]{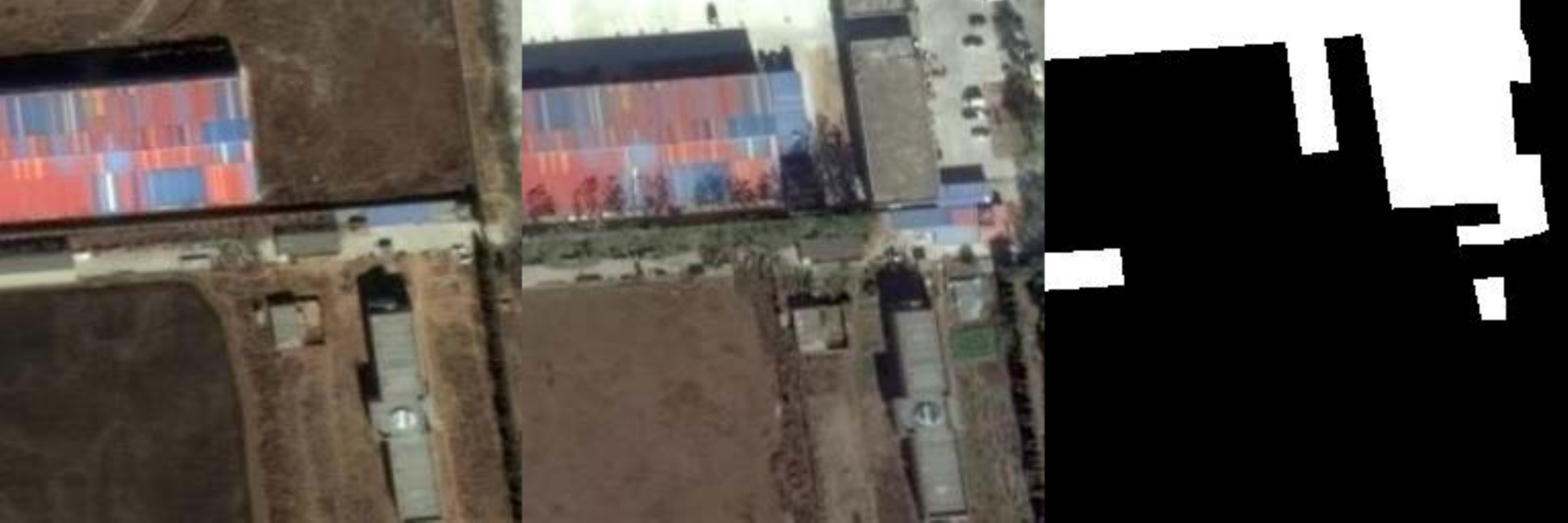}}
		\vspace{0.1cm}
	\end{minipage}
	\hfill
	\begin{minipage}[b]{0.2\linewidth}
		\centering
		\centerline{\includegraphics[width=4.4cm]{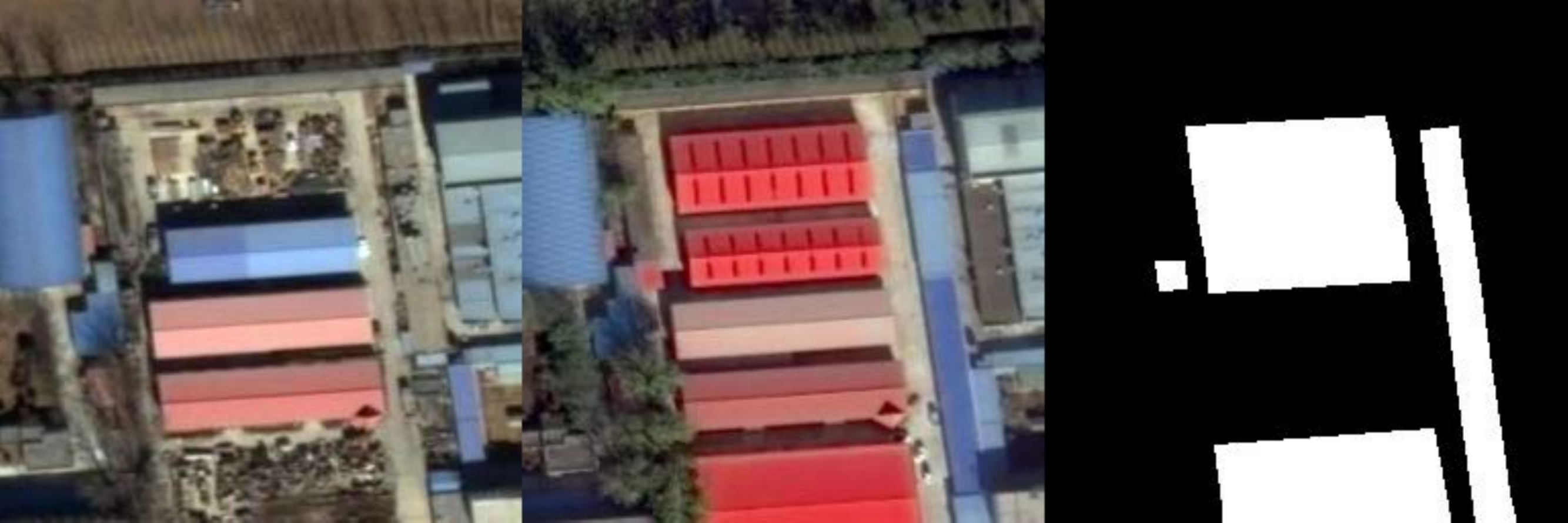}}
		\vspace{0.1cm}
	\end{minipage}
	\hfill
	\begin{minipage}[b]{0.2\linewidth}
		\centering
		\centerline{\includegraphics[width=4.4cm]{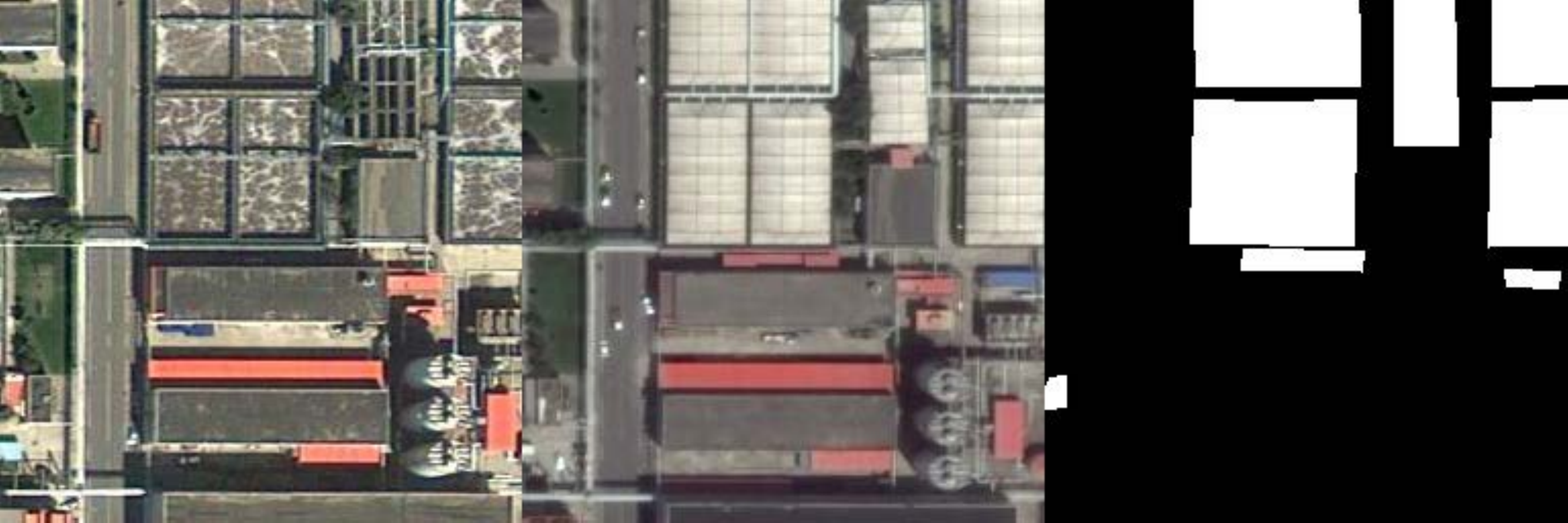}}
		\vspace{0.1cm}
	\end{minipage}
	\hfill
	\begin{minipage}[b]{0.2\linewidth}
		\centering
		\centerline{\includegraphics[width=4.4cm]{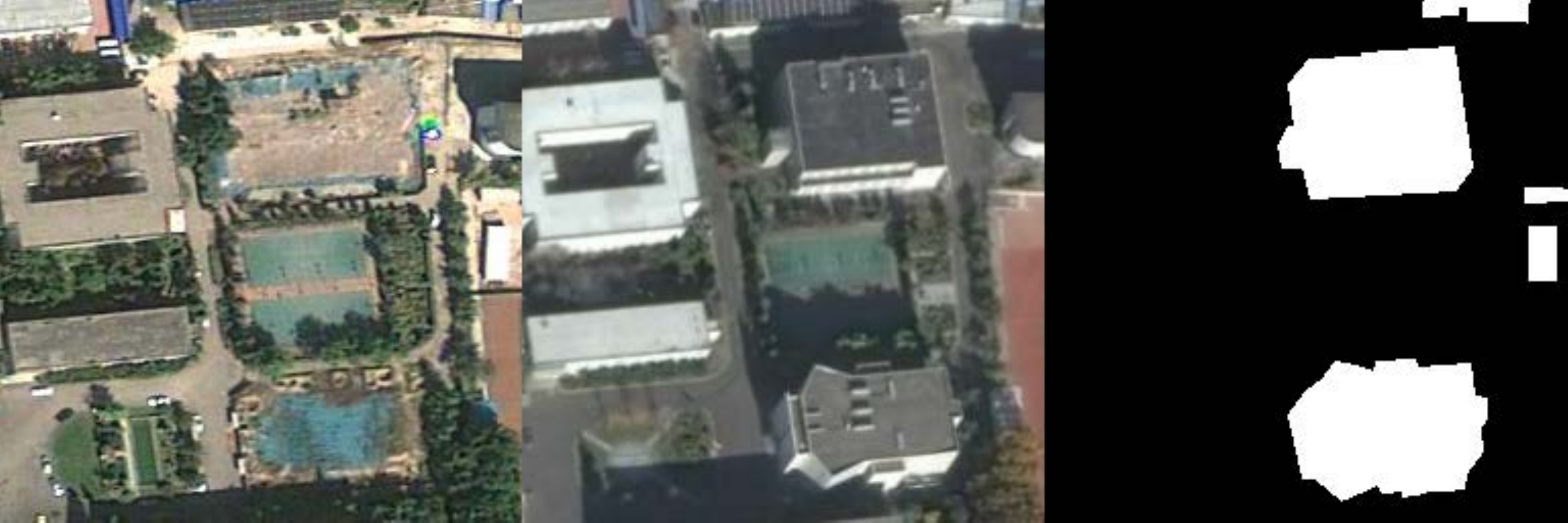}}
		\vspace{0.1cm}
	\end{minipage}
	\hspace{0.4cm}
	
	\vfill
	
	\hspace{0.4cm}
	\begin{minipage}[b]{0.2\linewidth}
		\centering
		\centerline{\includegraphics[width=4.4cm]{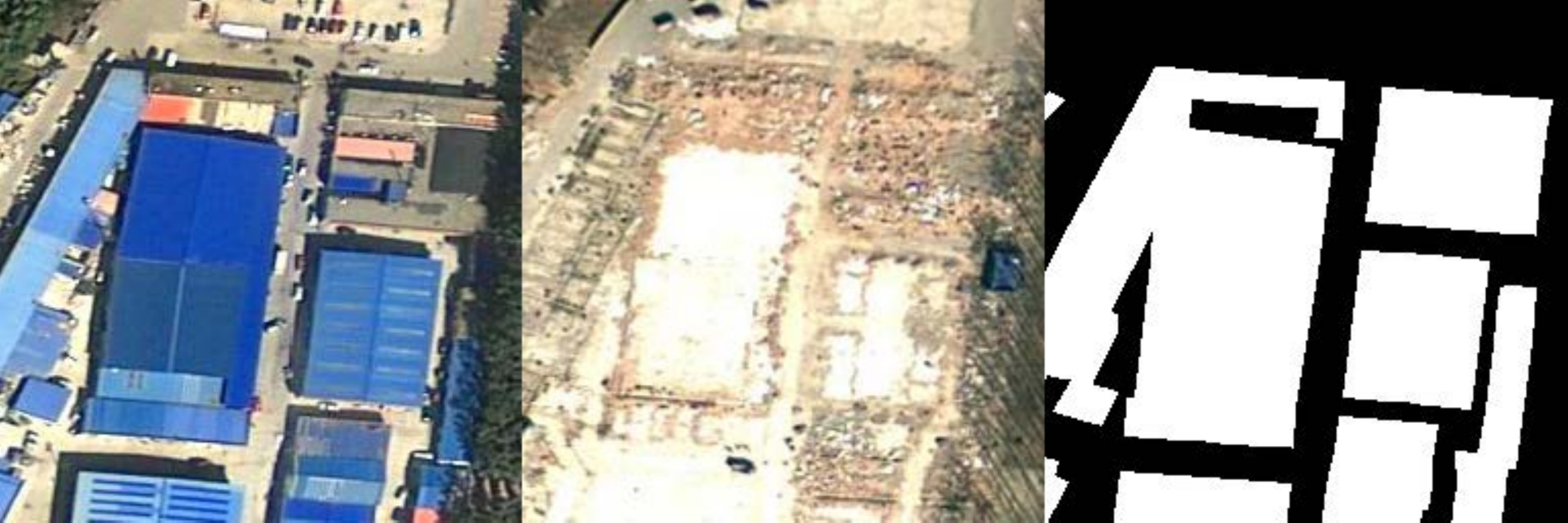}}
		\vspace{0.1cm}
	\end{minipage}
	\hfill
	\begin{minipage}[b]{0.2\linewidth}
		\centering
		\centerline{\includegraphics[width=4.4cm]{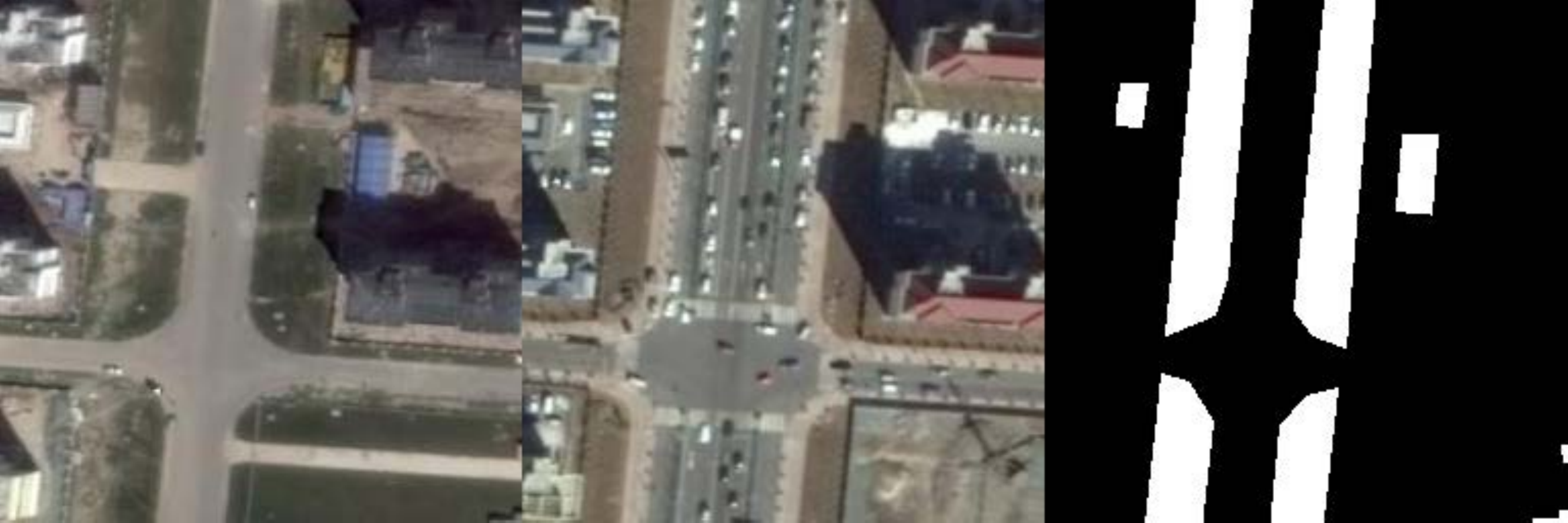}}
		\vspace{0.1cm}
	\end{minipage}
	\hfill
	\begin{minipage}[b]{0.2\linewidth}
		\centering
		\centerline{\includegraphics[width=4.4cm]{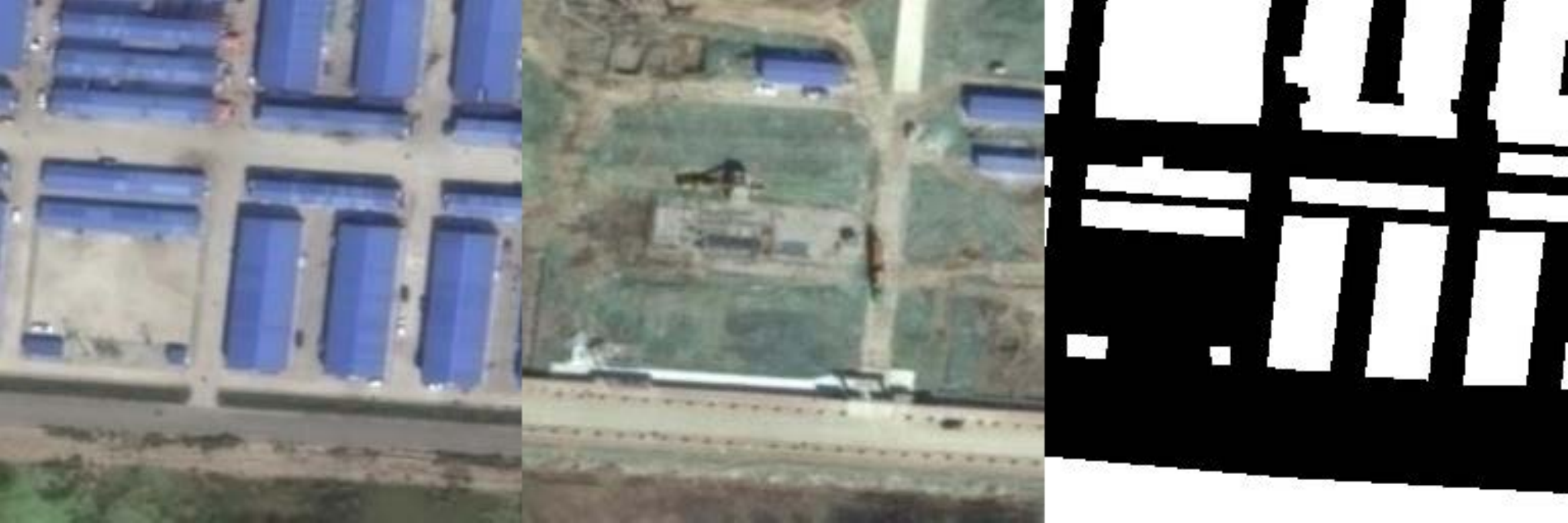}}
		\vspace{0.1cm}
	\end{minipage}
	\hfill
	\begin{minipage}[b]{0.2\linewidth}
		\centering
		\centerline{\includegraphics[width=4.4cm]{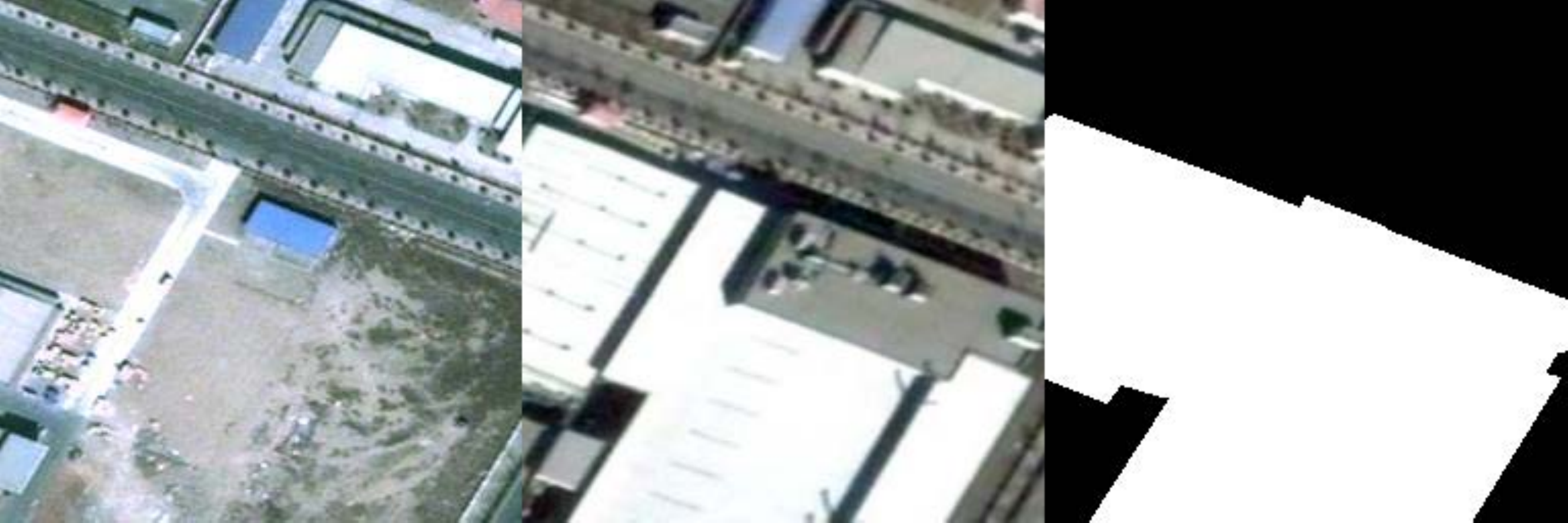}}
		\vspace{0.1cm}
	\end{minipage}
	\hspace{0.4cm}
	
	\vfill
	
	\hspace{0.4cm}
	\begin{minipage}[b]{0.2\linewidth}
		\centering
		\centerline{\includegraphics[width=4.4cm]{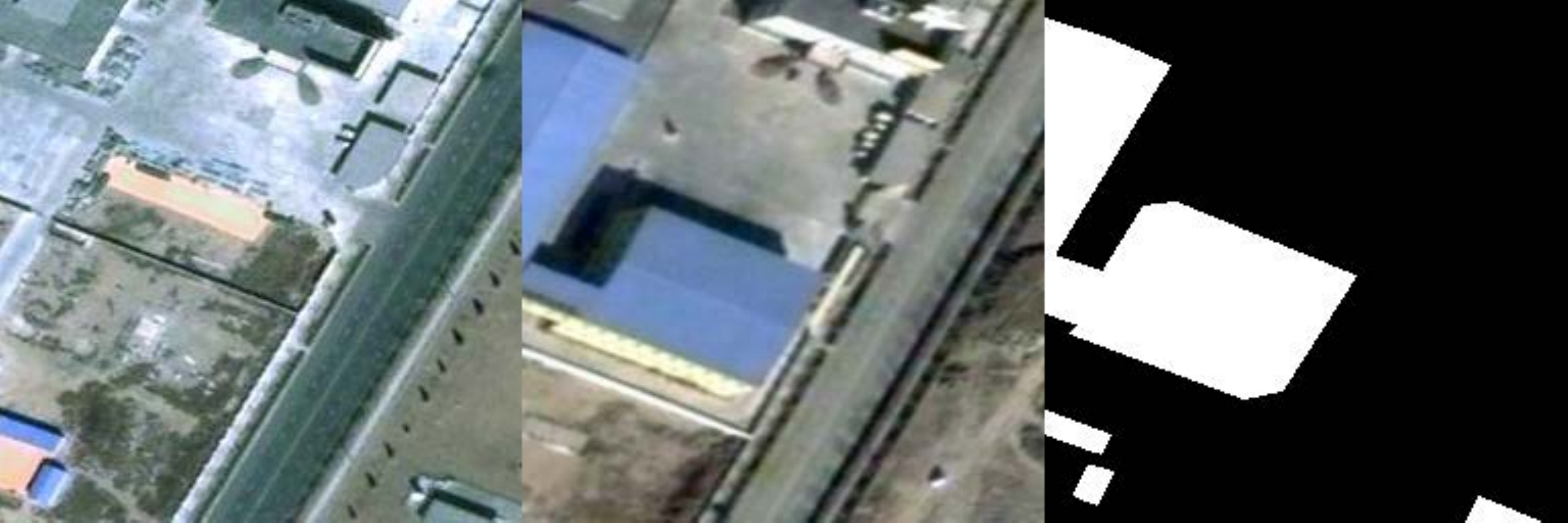}}
		\vspace{0.1cm}
	\end{minipage}
	\hfill
	\begin{minipage}[b]{0.2\linewidth}
		\centering
		\centerline{\includegraphics[width=4.4cm]{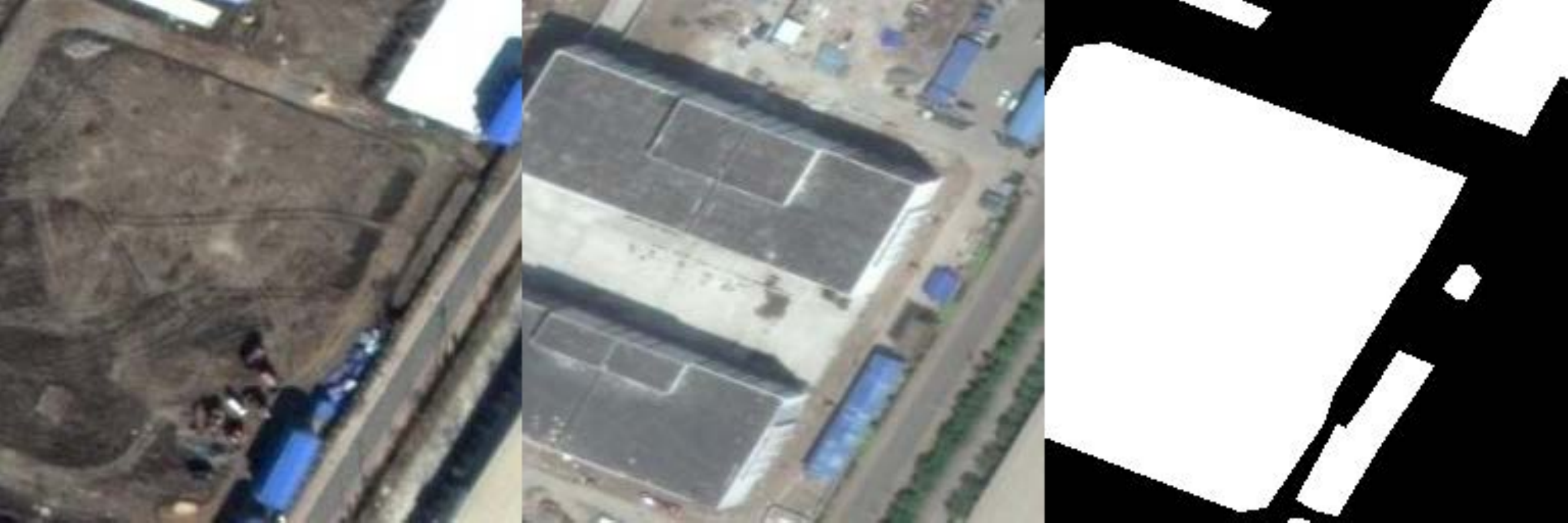}}
		\vspace{0.1cm}
	\end{minipage}
	\hfill
	\begin{minipage}[b]{0.2\linewidth}
		\centering
		\centerline{\includegraphics[width=4.4cm]{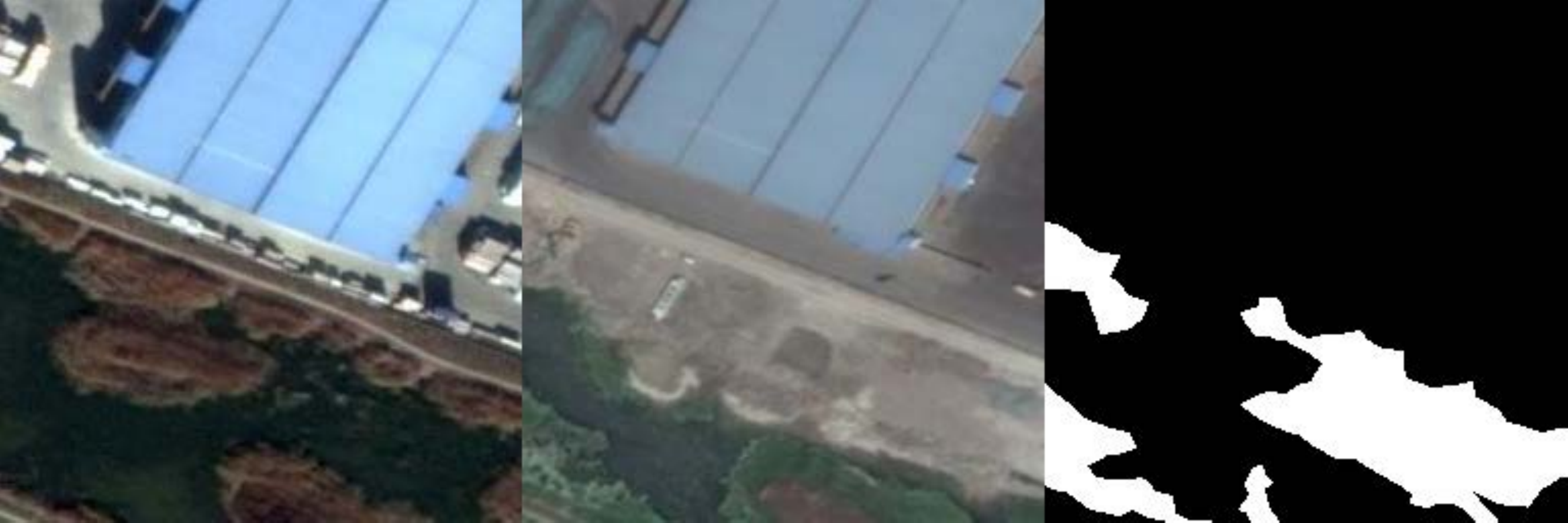}}
		\vspace{0.1cm}
	\end{minipage}
	\hfill
	\begin{minipage}[b]{0.2\linewidth}
		\centering
		\centerline{\includegraphics[width=4.4cm]{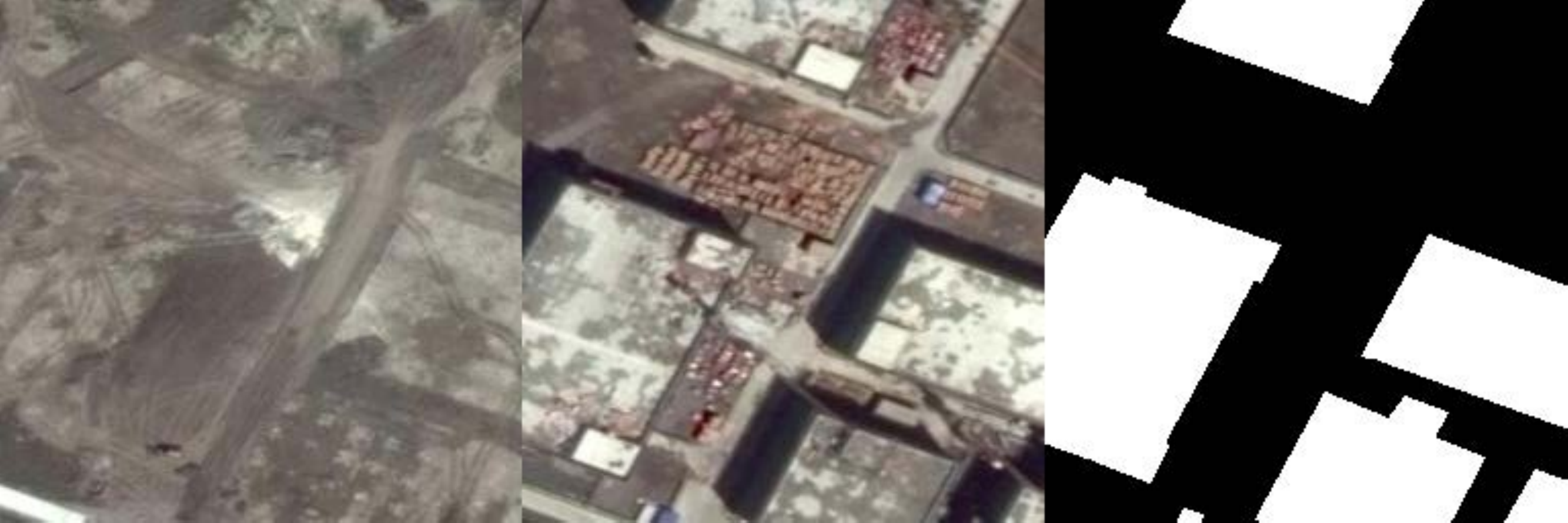}}
		\vspace{0.1cm}
	\end{minipage}
	\hspace{0.4cm}
	
	\vfill
	
	\hspace{0.4cm}
	\begin{minipage}[b]{0.2\linewidth}
		\centering
		\centerline{\includegraphics[width=4.4cm]{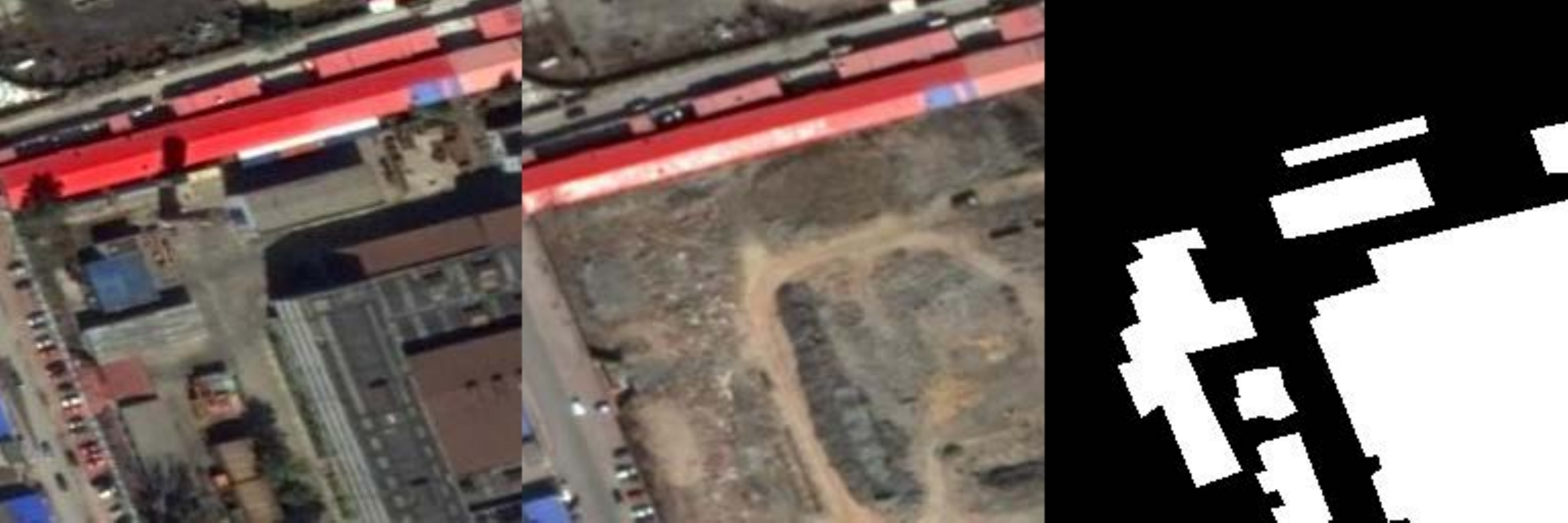}}
		\vspace{0.1cm}
	\end{minipage}
	\hfill
	\begin{minipage}[b]{0.2\linewidth}
		\centering
		\centerline{\includegraphics[width=4.4cm]{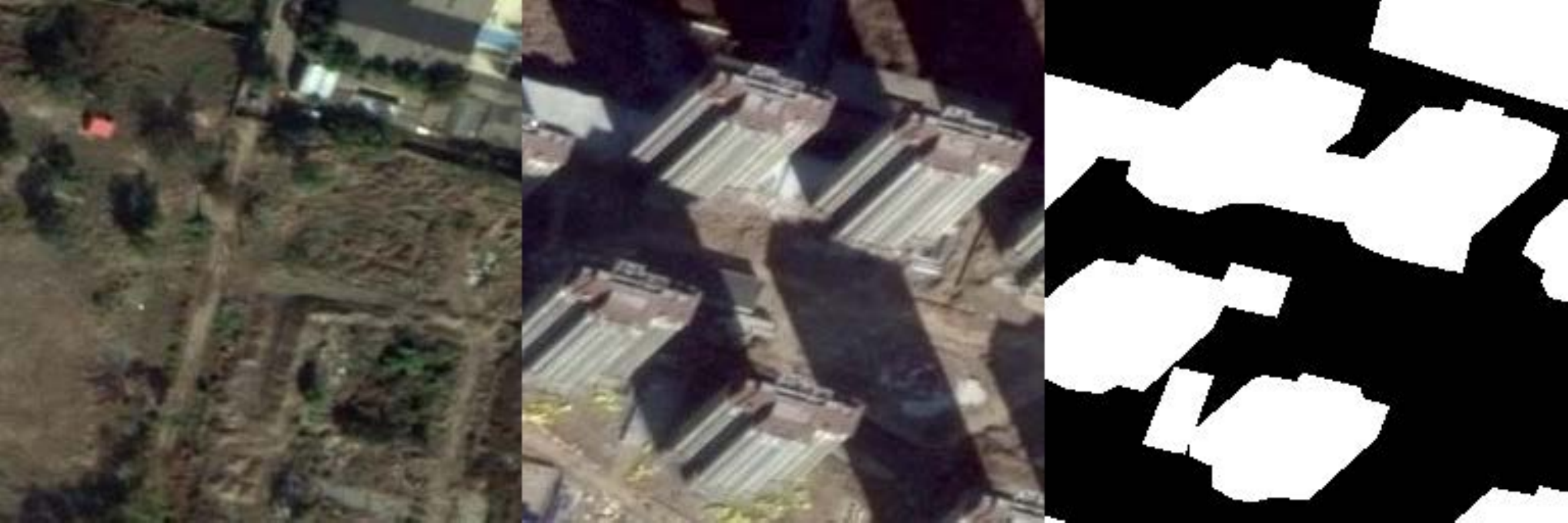}}
		\vspace{0.1cm}
	\end{minipage}
	\hfill
	\begin{minipage}[b]{0.2\linewidth}
		\centering
		\centerline{\includegraphics[width=4.4cm]{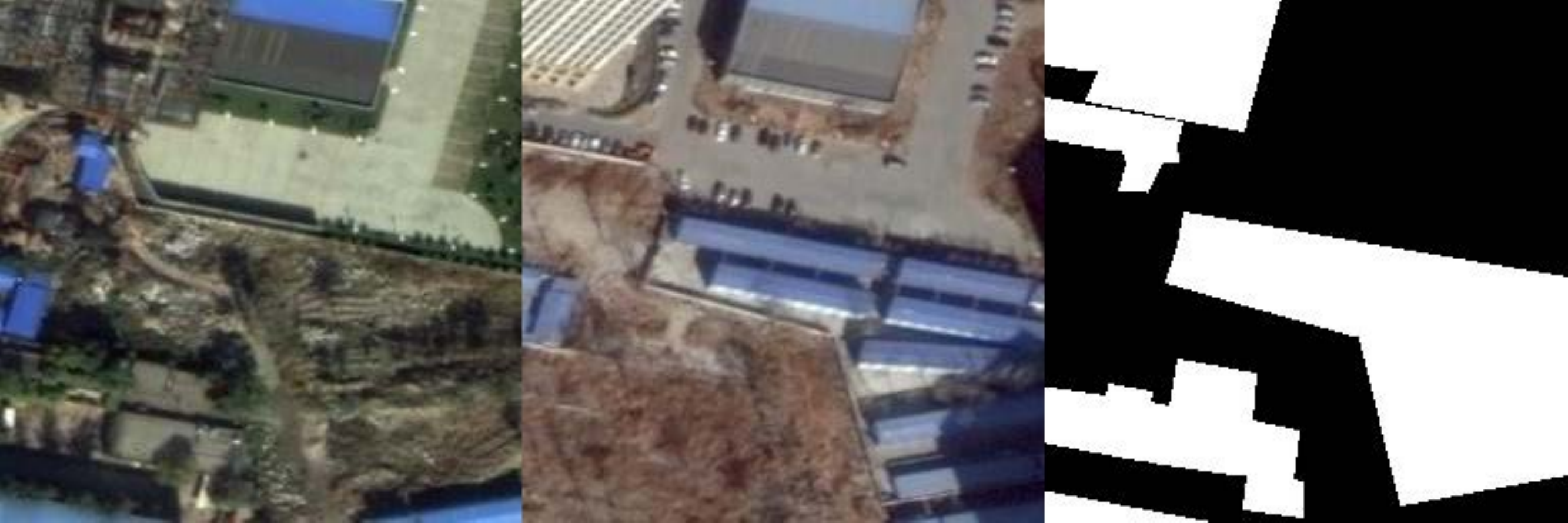}}
		\vspace{0.1cm}
	\end{minipage}
	\hfill
	\begin{minipage}[b]{0.2\linewidth}
		\centering
		\centerline{\includegraphics[width=4.4cm]{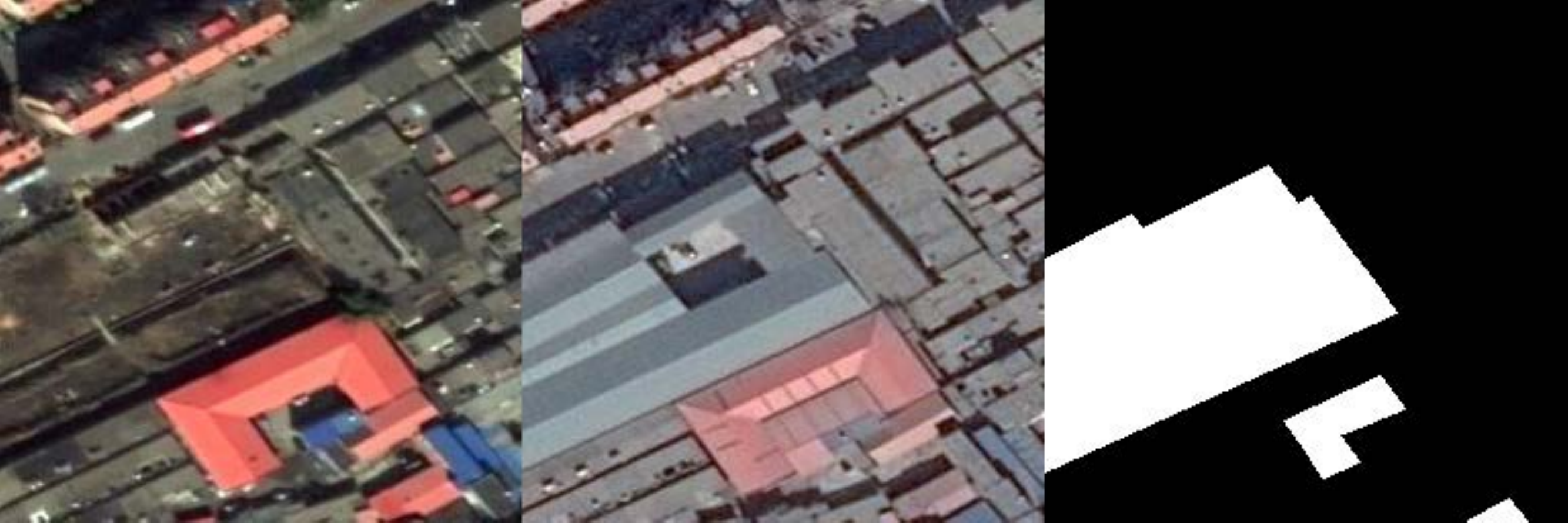}}
		\vspace{0.1cm}
	\end{minipage}
	\hspace{0.4cm}
	\vfill
	\caption{Image patch examples and corresponding reference images collected for our training.}
	\label{fig:exp-datasets-trainingdata}
	\vspace{-0.2cm}
\end{figure*}

\section{Experiments}\label{exp}
In this section, we first introduce the collection and annotation of training and testing data used in our experiments. Then we provide implementation details of our methods and evaluation metrics. Finally, we compare our proposed methods, i.e. W-Net and CDGAN, with well-established baselines to validate their effectiveness.
\subsection{Datasets}
To our best knowledge, currently there have been few available public datasets specifically for remote sensing change detection. To facilitate the training of our  proposed methods, as well as to benefit future research, we construct a large-scale dataset with high intra-class diversity and low inter-class dissimilarity for change detection, and provide ground truth annotations. For testing, we use two datasets, i.e. the Google Earth Dataset and the GF-2 Dataset, with details given below.     

\subsubsection{Data Collection}
For training data, we first collect 29 pairs of sample images covering two big cities, Beijing and Tianjin, in China with Google Earth during the period from 2006 to 2017,  with spatial resolution 0.46m (corresponding to 18 level google images).
Though Google Earth data are post-processed, there is no significant difference between Google Earth data and the real optical remote sensing images even in the pixel-level land use/cover mapping~\cite{Hu2013GoogleEarthData}. 
These images are captured at different times of a day and seasons under different imaging conditions, which increase the complexities and diversities of the data. They are with quite large sizes, e.g. $2000 \times 2000$ pixels, and are manually labeled (annotation details will be presented in Section~\ref{annotation}).  The classes of the objects in these captured images include buildings, waters, roads, bridges, residences, forests, farmland, etc., and mainly involve the changes caused by rapid urbanization. Buildings are the most significant component among these changes, thus in this work, we mainly focus on buildings.

Then, for each image of the 29 large image pairs, a set of $1,000$ image patches with size $256\times256$ is generated by randomly cropping from the original one. That makes a total of $29,000$ pairs of VHR optical remote sensing image patches, which are divided into 80\% for training and 20\% for validation. That is, we obtain a training set containing $23,200$ pairs of image patches and a validation set containing $5,800$ pairs of image patches.

To evaluate the performance of our proposed W-Net and CDGAN architectures, we also collect a testing dataset, which is named Google Earth. This dataset
consists of 13 image pairs all with a size of $500\times500$ pixels 
covering Beijing, China. They are captured with a spatial resolution of about 1m during the period of 2009 to 2015.
Besides this Google Earth dataset, we also use a GF-2 Dataset for testing. 
It contains 6 image pairs with size $500\times500$, spatial resolution about 1m for commercial use. These images are captured by GaoFen-2 satellite covering Beijing, China in 2005.

\subsubsection{Data Annotation}
\label{annotation}

We first explain the annotation of ground truths over the collected images for training. All the 29 pairs of large images are manually labelled using remote sensing software ENVI5.1 by five annotators with expertise in the remote sensing image processing. The labeling work takes about one month.
After annotation, we also invite three specialists to manually check the annotated labelled images, in order to ensure the annotation accuracy. The inspection work costs about two weeks. 
Some samples from our established training dataset are shown in Fig.~\ref{fig:exp-datasets-trainingdata}.

ENVI is a professional remote sensing processing software with multiple useful modules such as image registration, image pan-sharpening, etc. For labeling the images, the annotators first open the two images of a pair captured at different time and overlap them in ENVI. Then they apply the region of interest (ROI) module to select the change areas manually with polygons. In this process, we make sure the polygons to have enough edges to respect the boundaries of actual ground objects as much as possible. After that, all the labeled regions are converted to binary classification maps with only black and white colors, with black color denoting unchanged areas and white color denoting changed areas.

The testing data are annotated with ground truths following similar procedures for the training data.

\subsection{Implementation}
\label{subsec: gan-resnet}
We implement the proposed methods with TensorFlow framework~\cite{abadi2016tensorflow} and train them on a single GeForce GTX 1080Ti GPU using Adam optimizer~\cite{kingma2014adam}. During training, the networks are initialized by drawing weights from zero-mean Gaussian distributions with a standard deviation 0.02, and biases are initialized as 0. The learning rate is initially set as 0.0002 and reduced in an adaptive way. The momentum is 0.5. The training time for our W-Net and CDGAN is presented in Table~\ref{tb:exp-wnet-size}. To avoid the fast convergence of D network, G network is updated twice for each D network update. In testing, we crop image patches from the original images in a raster scan manner with a step of 128 pixels.
Then the outputs for each patch are stitched together to reconstruct the final results, where the pixel values in overlapping areas are averaged.

\begin{table}[htb]
	\tiny
	\centering
	\caption[justification=centering]{Model sizes and training time of W-Net and CDGAN. Note d stands for ``days'', h for ``hours'', and m for ``minutes''.}
	\resizebox{0.45\textwidth}{!}{%
		\begin{tabular}{crrc}
			\Xhline{0.5pt}
			\rowcolor{mygray}
			Method                  & Parameter Number & Training Time & Batch Size \\
			\hline

			W-Net					&  42,570,625        & 1d-15h-12m       & 22         \\
			
			CDGAN                   & 123,045,378    & 6d-18h-54m      & 18         \\
			\Xhline{0.5pt}
	\end{tabular}}%
	\label{tb:exp-wnet-size}
\end{table}

\subsection{Evaluation Metrics}\label{exp-criteria}

Both qualitative and quantitative analysis are employed to evaluate performance of the proposed  methods. For quantitative evaluation, missed alarm rate (MAR), false alarm rate (FAR), overall error rate (OER), and kappa coefficient ($kappa$) are adopted as metrics. The first three metrics are computed respectively by $MAR=FN/(TP+FN)$,  $FAR=FP/(FP+TN)$,  $OER=(FP+FN)/(TP+TN+FP+FN)$, in which false negative (FN) denotes the number of pixels wrongly classified as unchanged ones; false positive (FP) represents the number of pixels wrongly classified as the changed ones, true negative (TN) means the number of pixels correctly classified as unchanged ones, and true positive (TP) is the number of pixels correctly classified as the changed ones. $kappa$ is usually used for measuring classification performance and a higher $kappa$ value means better performance. It is calculated as 
\begin{equation}
\begin{split}
kappa&=\frac{PCC-PRE}{1-PRE},\\
\text{}\ PCC&=\frac{TP+TN}{TP+TN+FP+FN},\\
\text{}\ PRE&=\frac{(TP+FN)\cdot(TP+FP)}{(TP+TN+FP+FN)^2}\\
&+\frac{(TN+FP)\cdot(TN+FN)}{(TP+TN+FP+FN)^2},
\end{split}
\end{equation}
where $PCC$ denotes the percentage of correct classifications, and $PRE$ represents that of expected agreements. 

For qualitative evaluation, we provide FAR-MAR (FM) curves, which are typically considered as a global detection performance indicator and  measure the agreement between detection results and reference images. The FM curves for good results should be close to the bottom-left corner of the coordinate system where horizontal axis indicates MAR and vertical axis indicates FAR. A smaller area under FM curves (AUC) typically indicates higher quality of the change map. Besides, precision and recall (PR) curves are also used. Precision and recall can be defined as $Precision=TP/(TP+FP)$ and $Recall=TP/(TP+FN)$, respectively. Correspondingly, they should be close to the top-right corner of the coordinate system with recall as the horizontal axis and precision as the vertical.  

\subsection{Comparison of W-Net and CDGAN}

\begin{figure*}[!ht]
	\centerline{\includegraphics[width=1.\textwidth]{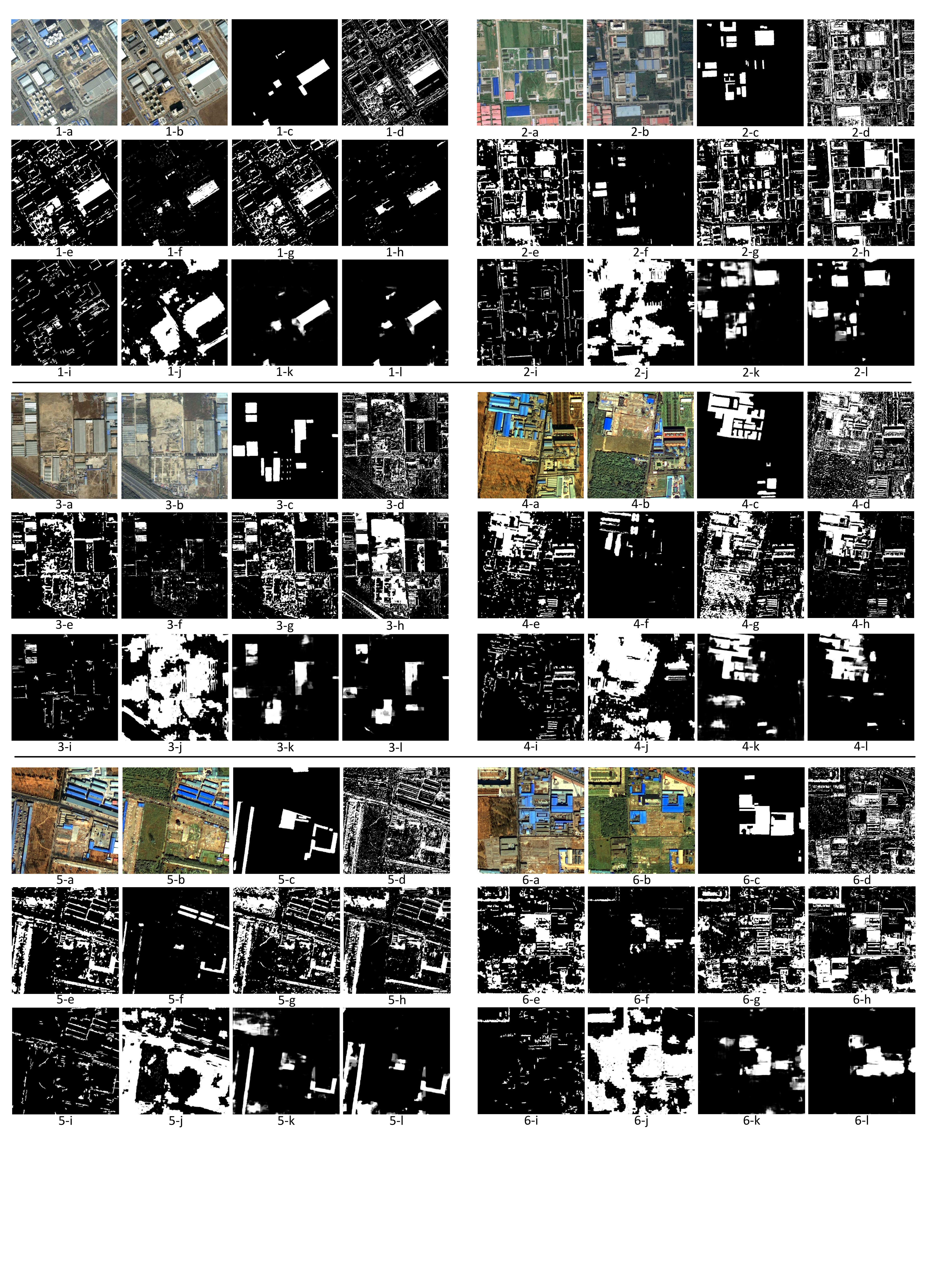}}
	\caption{Change detection results of the proposed W-Net and CDGAN,  and other seven baseline methods. The 1st, 2nd and 3rd  image pairs are from Google Earth. The 4th, 5th and 6th image pairs are from GF-2. (a) Image at t1. (b) Image at t2. (c) Ground Truth. (d) EM. (e) MRF. (f) IR-MAD. (g) PCA. (h) Parcel. (i) MBI. (j) SHC. (k) W-Net. (l) CDGAN.}
	\label{fig:exp-wnet-gan}
\end{figure*}

\begin{table*}
	\tiny
	\centering
	\caption[justification=centering]{Performance comparisons of different approaches on Google Earth dataset and GF-2 dataset.}

	\resizebox{\textwidth}{!}{ %
	\begin{tabular}{c cccc cccc}
		\Xhline{0.5pt}
		\toprule
		\multirow{2}*{Methods} & \multicolumn{4}{c}{GoogleEarth-Dataset} & \multicolumn{4}{c}{GF-Dataset} \\
		\cmidrule(lr){2-5} \cmidrule(lr){6-9} 
		& FAR & MAR & OER & $kappa$ & FAR & MAR & OER & $kappa$ \\
		\midrule	
		EM-based~\cite{bruzzone2000automatic}    
		& 0.3103 & 0.3367 & 0.3174 & 0.1184
		& 0.3016 & 0.4503 & 0.3239 & 0.1567\\
		MRF-based~\cite{bruzzone2000automatic}   
		& 0.2630 & 0.2771 & 0.2703 & 0.1690
		& 0.1936 & 0.4280 & 0.2296 & 0.2860\\
		IR-MAD-based~\cite{nielsen2007regularized}
		& 0.0252 & 0.4543 & 0.0642 & 0.4332
		& 0.0160 & 0.7425 & 0.1223 & 0.3276\\
		PCA-based~\cite{Celik2009Unsupervised}  
		& 0.2231 & 0.3832 & 0.2393 & 0.1548
		& 0.2547 & 0.5311 & 0.2957 & 0.1540\\
		Parcel-based~\cite{bovolo2009multilevel}
		& 0.2382 & 0.4418 & 0.2519 & 0.1509
		& 0.1934 & 0.4211 & 0.2266 & 0.3028\\
		MBI-based~\cite{Huang2013Building}
		& 0.0396 & 0.7898 & 0.0913 & 0.1425
		& 0.0483 & 0.8714 & 0.1675 & 0.1056\\
		SHC-based~\cite{Ding2015Sparse}   
		& 0.3805 & 0.0968 & 0.3657 & 0.1353
		& 0.4184 & 0.1034 & 0.3729 & 0.2466\\
		W-Net    
		& 0.0276 & 0.2949 & 0.0463 & 0.6088
		& 0.0404 & 0.5008 & 0.1101 & 0.5015\\
		CDGAN    
		& 0.0195 & 0.3029 & \textbf{0.0409} & \textbf{0.6411}
		& 0.0233 & 0.5291 & \textbf{0.1012} & \textbf{0.5053}\\
		
		\bottomrule
		\Xhline{0.5pt}
		\vspace{-0.3cm}
	\end{tabular}}%
	\label{tab:exp-wnet-gan}
\end{table*}

\begin{figure*}
	\centering
	\subfloat[]{\includegraphics[width = 0.45\linewidth]{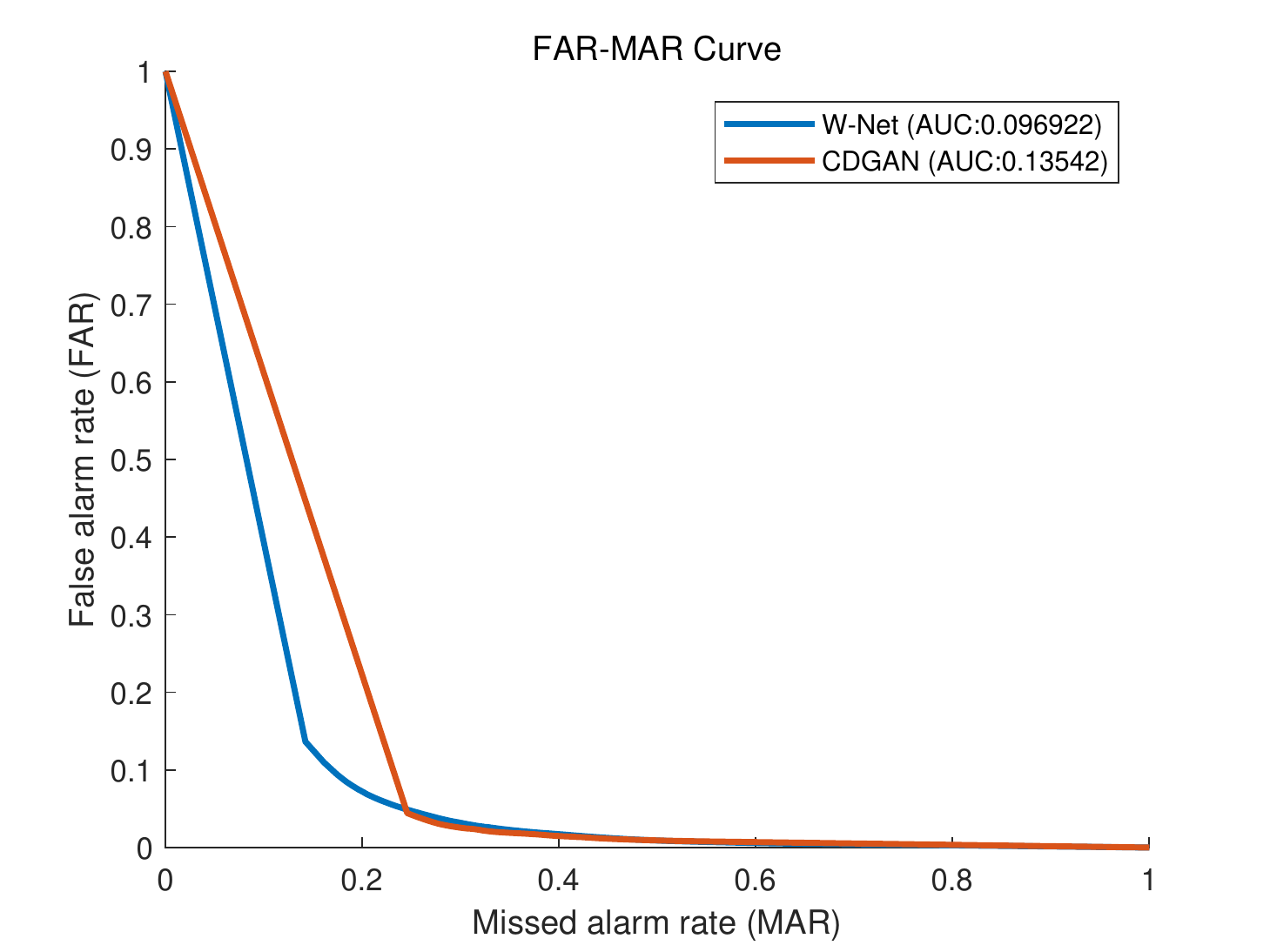}
		\label{fig:GoogleEarth-FM2}		}
	\subfloat[]{\includegraphics[width = 0.45\linewidth]{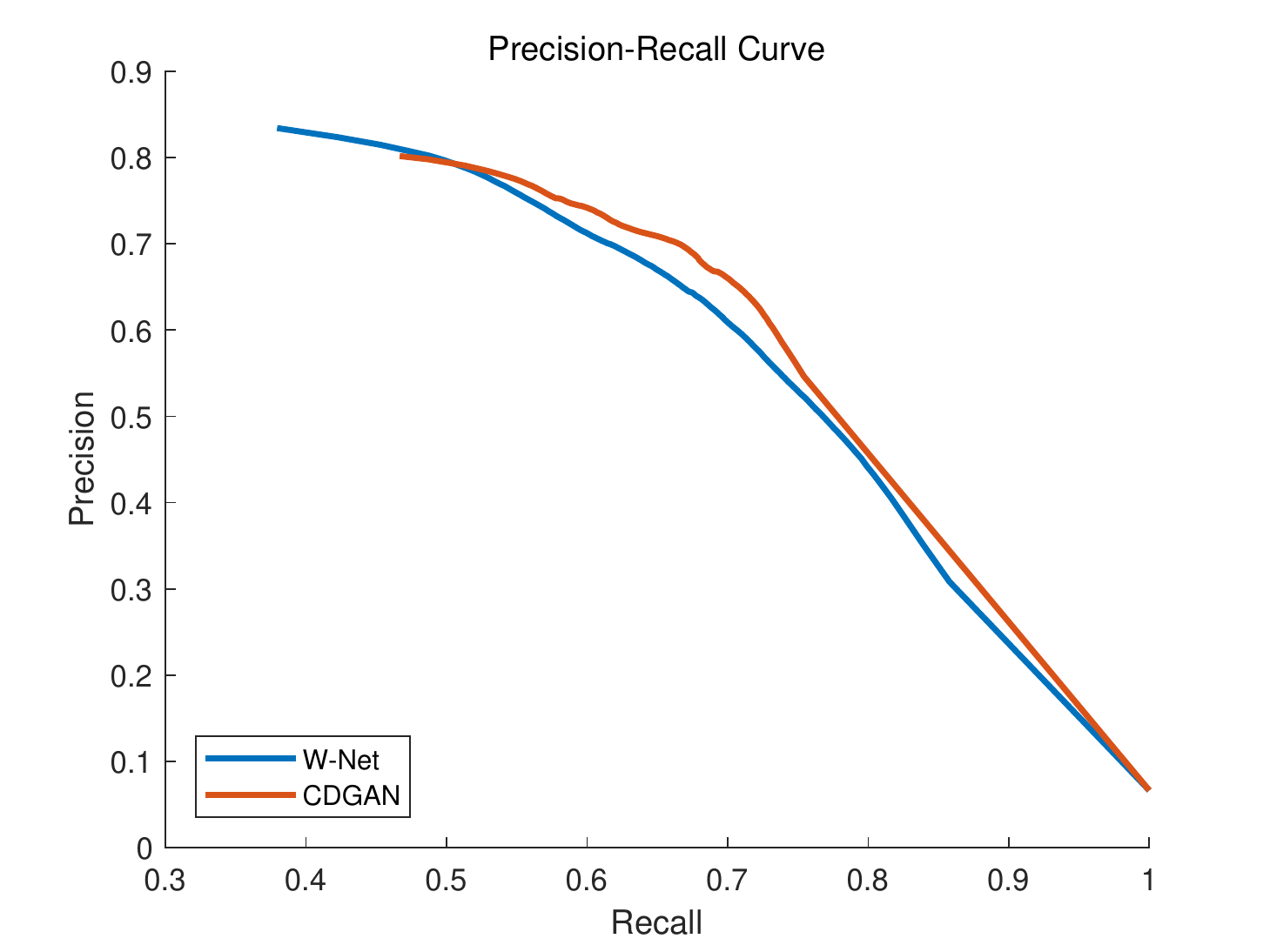}
		\label{fig:GoogleEarth-PR2}		}\vfill
	\subfloat[]{\includegraphics[width = 0.45\linewidth]{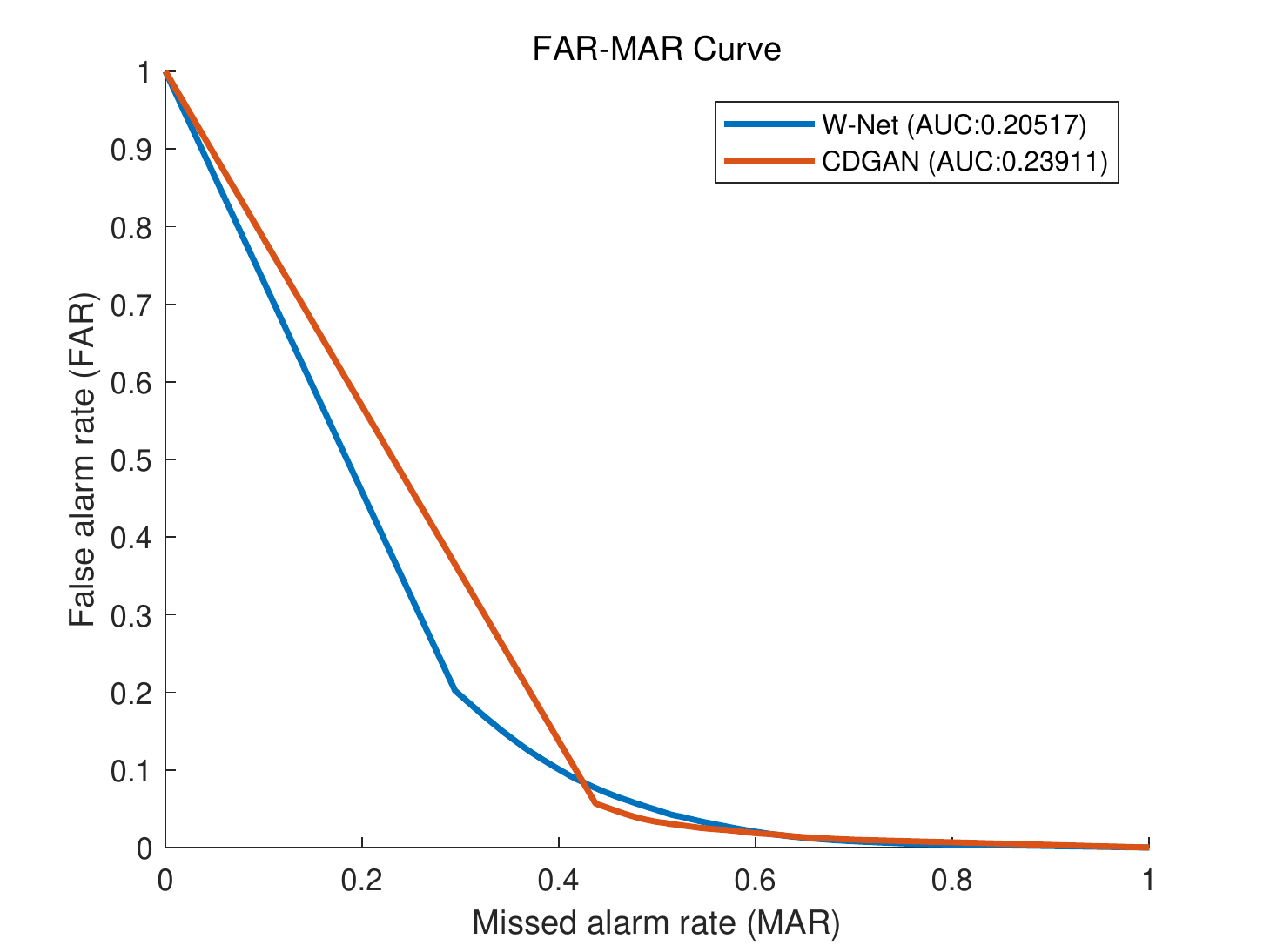}
		\label{fig:GF-FM2}		}
	\subfloat[]{\includegraphics[width = 0.45\linewidth]{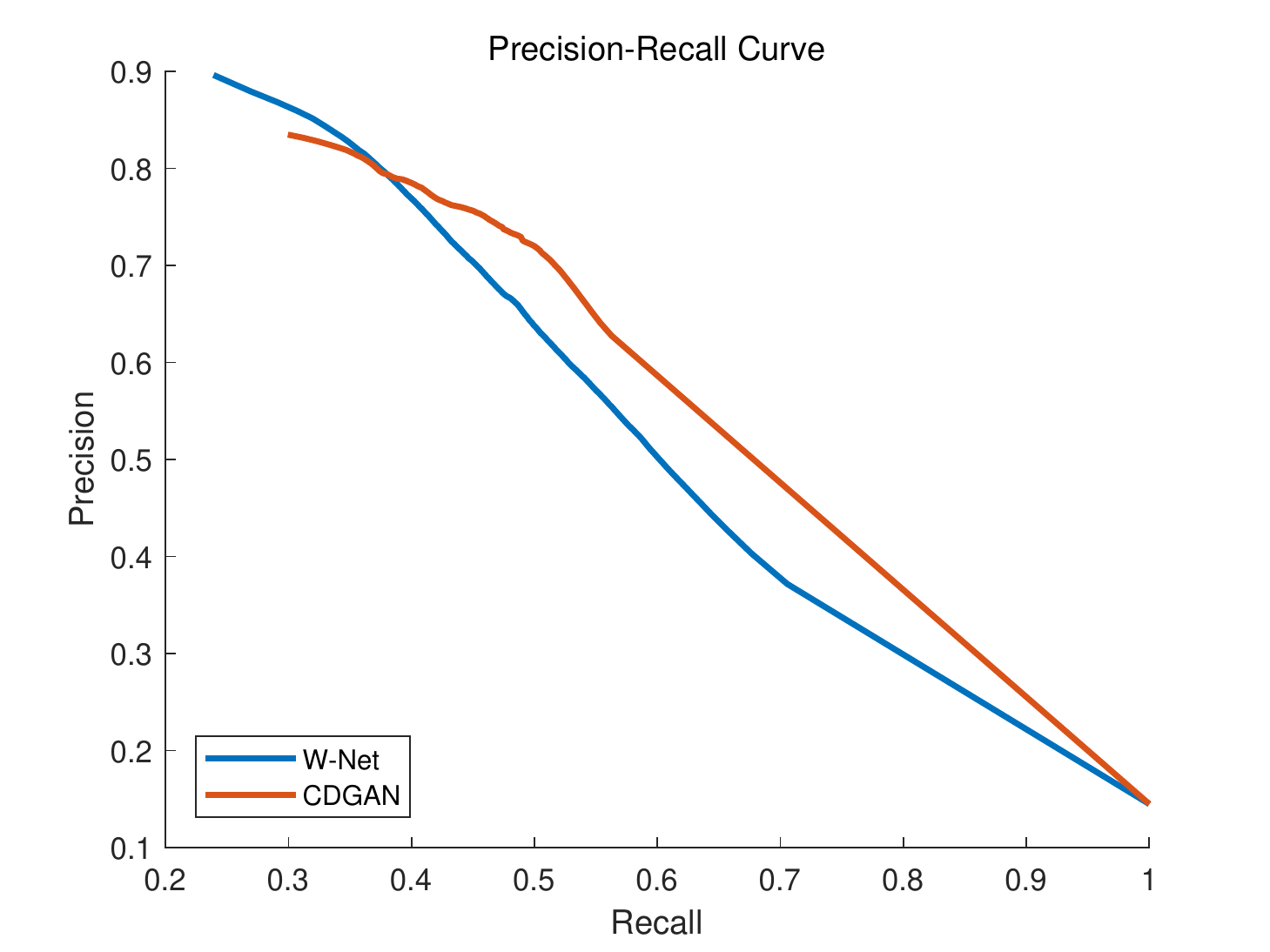}
		\label{fig:GF-PR2}		}
	\caption{Quantitative analysis on FAR, MAR, precision and recall of two different network architectures on Google Earth dataset and GF-2 dataset, i.e. W-Net and CDGAN. (a) FM curve on Google Earth dataset. (b) PR curve on Google Earth dataset. (c) FM curve on GF-2 dataset. (d) PR curve on GF-2 dataset.}
	\label{fig:exp-wnet-gan-curve}
\end{figure*}

We first compare the performance of W-Net and CDGAN for change detection on Google Earth dataset and GF-2 dataset. We list all quantitative results in terms of the four metrics including FAR, MAR, OER and $kappa$ in Table~\ref{tab:exp-wnet-gan}. Some example image pairs from these two datasets and corresponding results are shown in Fig.~\ref{fig:exp-wnet-gan} (k) (l). We also plot FM curves and PR curves for each architecture in Fig. \ref{fig:exp-wnet-gan-curve}. 

On both testing datasets, i.e. Google Earth and GF-2, CDGAN performs better than W-Net w.r.t. all the metrics except MAR. CDGAN gains slightly higher MAR than W-Net, but the difference is almost negligible. Fig. \ref{fig:exp-wnet-gan-curve} illustrates the FM and PR curves. According to FM curves, W-Net is superior to CDGAN with a smaller AUC area. However, as for PR curves, CDGAN is more preferable because it is basically above W-Net and closer to the top right corner in the plotted figure. These results indicate that CDGAN is generally better than W-Net.

Comparing their quantitative performance from the perspective of testing data, GF-2 dataset has more complex spectral distributions than Google Earth dataset. The overperfect learning of strided convolution on Google Earth may affect their  performance on GF-2 dataset to some extent. This may be the reason why W-Net and CDGAN achieve better performance regarding the four metrics on Google Earth dataset than on GF-2 dataset.

For qualitative comparison, as shown in Fig. \ref{fig:exp-wnet-gan} (k) (l), the proposed two methods do not show much difference w.r.t. missed detection. Both methods do not miss much changed area compared with ground truth. CDGAN has better performance with fewer false detection than W-Net. 
From the second, fourth and fifth image pairs in Fig.~\ref{fig:exp-wnet-gan}, it can be seen that W-Net generates many false detections, while CDGAN successfully gets rid of these pseudo changes and generates more accurate and clearer results. 
Moreover, W-Net blurs some locations belonging to the changed class. Nevertheless, the performance of CDGAN is better over the main detection regions.

Through the above comparisons, we can see that
generally, CDGAN obtains better experimental results than W-Net.  
GAN can learn more substantial distributions and relieves information loss in training data, consequently leading to better transferability and generalization.

\subsection{Comparison of Proposed Methods with Other Baselines}\label{exp-methods}

To test the effectiveness of the proposed W-Net and CDGAN, we use seven approaches as baselines to make comparisons. 
The baselines methods include:

1) EM-based method \cite{bruzzone2000automatic} (Fig. \ref{fig:exp-wnet-gan} (d)); 

2) MRF-based method \cite{bruzzone2000automatic} (Fig. \ref{fig:exp-wnet-gan} (e));

3) IR-MAD-based method \cite{nielsen2007regularized} (Fig. \ref{fig:exp-wnet-gan} (f));

4) PCA-based method \cite{Celik2009Unsupervised} (Fig. \ref{fig:exp-wnet-gan} (g));

5) Parcel-based method \cite{bovolo2009multilevel} (Fig. \ref{fig:exp-wnet-gan} (h));

6) Morphological building index (MBI)-based method \cite{Huang2013Building} (Fig. \ref{fig:exp-wnet-gan} (i));

7) Sparse hierarchical clustering (SHC)-based method \cite{Ding2015Sparse} (Fig. \ref{fig:exp-wnet-gan} (j)).

The involved parameters of these methods are set as those in their original papers. The experiments are conducted also on two testing datasets Google Earth and GF-2. The quantitative comparisons are presented in Table~\ref{tab:exp-wnet-gan}, and the qualitative results are summarized in Fig.~\ref{fig:exp-wnet-gan}. 

For quantitative comparison, it can be seem from Table~\ref{tab:exp-wnet-gan}, CDGAN
outperforms all of the compared methods in terms of OER and $kappa$ metrics, and our W-Net is the second best, on both of the testing datasets. Regarding the FAR metric, our W-Net and CDGAN are among the best methods on both Gooogle Earth and GF-2 testing datasets. For example, CDGAN ranks the first among all the methods in terms of FAR on Google Earth, and the second place on GF-2; W-Net is the third best method on both datasets. However, we also observe that our methods do not achieve competitive MAR performance. This can be possibly explained as follows. First, our training data size is rather limited considering the data-hunger of CNNs; second, we do not use data augmentation; third, we use random initialization and do not use pretrained backbone. From an overall perspective, the results validate that our proposed W-Net and CDGAN are able to work well for change detection tasks. They can learn the feature differences of the changed pixel-pairs and shrink the feature differences between the unchanged pixel-pairs. Also, they can learn the intrinsic connections between the inputs and ground truth through training the conv-deconv networks.

Fig.~\ref{fig:exp-wnet-gan} shows some change maps obtained by all the seven baseline methods and our proposed W-Net and CDGAN. It is obvious that the change maps generated by our methods are very close to the ground truths. The results of the baseline methods look relatively messy and discontinuous, while our methods generate clearer unchanged regions and more accurate boundaries
for changed regions. Especially, compared with a large number of missed and false alarms of other methods, our methods  achieve the best results, as shown in the third image pair of Fig. 4, which is rather challenging. In addition, our methods avoid pseudo vegetation changes caused by seasons as shown in the second Google Earth image and all the GF-2 images of Fig. 4.
Our proposed methods are robust to noise, due to the integration of spectral and spatial information. Deep network architectures construct more powerful and discriminative feature representation. It is demonstrated that the change information extraction and classification through network learning may be helpful to mining the substantial difference information and avoiding redundant distracting information.

From the above quantitative and qualitative comparisons, it can be verified that the proposed W-Net and CDGAN are able to achieve more satisfactory results than the state-of-the-art change detection baselines. Our proposed methods not only have better performance on homologous data to training data, but also perform similarly well on heterogenous data, i.e. GF-2 dataset, showing  good generalization ability. This could be owing to the conv-deconv architectures and the introduction of GAN. In addition, strided convolutions, concatenation strategy and short connection also reinforce the feature learning ability and noise immunity of the proposed networks. The proposed end-to-end networks can adaptively learn the latent distributions, which is implemented by two steps of generation and classification of difference images as in traditional methods.

\vspace{-0.1cm}
\section{Conclusion}\label{conclusion}

In this paper, we propose to apply deep learning techniques to tackling change detection on high resolution remote sensing images. Firstly, we construct a dual-branch W-Net architecture, which adopts strided convolution, concatenation and short connection strategies to realize change features extraction and classification. Then, a GAN is designed, which adopts our W-Net as generator to learn a mapping function that reveals the distributions.  Two datasets,  Google Earth dataset and GF-2 dataset are used to verify their effectiveness through both qualitative and quantitative comparisons. The experiments show that our W-Net and CDGAN obtain effective results, and CDGAN is able to bring some performance improvement over W-Net. 
Different from the traditional methods, our methods can obtain final change maps directly from the two original images. In the future, we will attempt to devise better network architectures based on our current work for change detection. In addition, we also try to transfer our models by using GF-2 or other datasets for training.

\ifCLASSOPTIONcaptionsoff
  \newpage
\fi

\bibliographystyle{IEEEtran}
\bibliography{myIEEERef}
\end{document}